\documentclass[10pt,journal,compsoc]{IEEEtran}
\usepackage{amsmath,amsfonts}

\usepackage{array}
\usepackage[caption=false,font=normalsize,labelfont=sf,textfont=sf]{subfig}
\usepackage{textcomp}
\usepackage{stfloats}
\usepackage{url}
\usepackage{verbatim}
\usepackage{graphicx}
\usepackage{cite}
\usepackage{multirow}
\usepackage{booktabs}

\usepackage{color}
\usepackage{hyperref}
\usepackage{graphicx}
\usepackage{amsmath}
\usepackage{amssymb}
\usepackage{cleveref}
\usepackage[section]{placeins}
\usepackage{float}
\usepackage{diagbox}

\usepackage{epstopdf}
\usepackage{amsmath}
\usepackage{amssymb}
\usepackage{multirow}
\usepackage{hyperref}
\usepackage{color}
\usepackage [latin1]{inputenc}
\usepackage{amsthm,amsmath,amssymb}
\usepackage{mathrsfs}
\usepackage{algorithm}
\usepackage{algorithmic}

\usepackage{cite}

\ifCLASSINFOpdf
\else
\fi
\begin{document}
%
\title{Meta Invariance Defense Towards Generalizable Robustness to Unknown Adversarial Attacks}
%
%
%
%

\author{Lei~Zhang,
        Yuhang Zhou,
        ~Yi Yang,
        Xinbo Gao
\IEEEcompsocitemizethanks{\IEEEcompsocthanksitem This work was partially supported by National Key R\&D Program of China (2021YFB3100800), National Natural Science Fund of China (62271090), Chongqing Natural Science Fund (cstc2021jcyj-jqX0023), and National Youth Talent Program. 

\IEEEcompsocthanksitem Lei Zhang is with the School of Microelectronics and Communication Engineering, Chongqing University, Chongqing 400044, China, and Peng Cheng Lab, Shenzhen, China.
(E-mail: leizhang@cqu.edu.cn)

\IEEEcompsocthanksitem Yuhang Zhou is with the School of Microelectronics and Communication Engineering, Chongqing University, Chongqing 400044, China.
(E-mail: yuhangzhou@cqu.edu.cn)

\IEEEcompsocthanksitem Yi Yang is with the College of Computer Science and Technology, Zhejiang University, Hangzhou 310058, China. (E-mail: yangyics@zju.edu.cn).

\IEEEcompsocthanksitem Xinbo Gao is with the Chongqing Key Laboratory of Image Cognition, Chongqing University of Posts and Telecommunications, Chongqing 400065, China. (E-mail: gaoxb@cqupt.edu.cn)}

\thanks{Manuscript received April 19, 2005; revised August 26, 2015.}}

%
%

\markboth{IEEE Transactions on Pattern Analysis and Machine Intelligence}%
{Shell \MakeLowercase{\textit{et al.}}: Bare Demo of IEEEtran.cls for Computer Society Journals}
%



\IEEEtitleabstractindextext{%
\begin{abstract}
Despite providing high-performance solutions for computer vision tasks, the deep neural network (DNN) model has been proved to be extremely vulnerable to adversarial attacks. Current defense mainly focuses on the known attacks, but the adversarial robustness to the unknown attacks is seriously overlooked. Besides, commonly used adaptive learning and fine-tuning technique is unsuitable for adversarial defense since it is essentially a zero-shot problem when deployed. Thus, to tackle this challenge, we propose an attack-agnostic defense method named \textbf{M}eta \textbf{I}nvariance \textbf{D}efense (MID). Specifically, various combinations of adversarial attacks are randomly sampled from a manually constructed \emph{Attacker Pool} to constitute different defense tasks against unknown attacks, in which a student encoder is supervised by multi-consistency distillation to learn the attack-invariant features via a meta principle. The proposed MID has two merits: 1) Full distillation from pixel-, feature- and prediction-level between benign and adversarial samples facilitates the discovery of attack-invariance. 2) The model simultaneously achieves robustness to the imperceptible adversarial perturbations in high-level image classification and attack-suppression in low-level robust image regeneration. Theoretical and empirical studies on numerous benchmarks such as ImageNet verify the generalizable robustness and superiority of MID under various attacks.
\end{abstract}

\begin{IEEEkeywords}
Meta-Defense, Attack-Invariant Features, Adversarial Attack, Generalizable Robustness, Deep Neural Network.
\end{IEEEkeywords}}

\maketitle

\IEEEdisplaynontitleabstractindextext

%
\IEEEpeerreviewmaketitle

\IEEEraisesectionheading{\section{Introduction}\label{sec:introduction}}
\IEEEPARstart{D}{eep} Neural Network (DNN) brings a high-performance solution for computer vision (CV) tasks. However, \cite{szegedy2013intriguing} proposes that DNN is extremely vulnerable to adversarial perturbations. That is, manually adding some carefully designed but human imperceptible noise to the input images can easily fool the DNN model to make totally wrong prediction. This is the so-called adversarial attack, which poses a great threat to the actual deployment of the DNN-based systems. For example, criminals are likely to escape from the DNN-based monitoring system through well-designed perturbations. Given the non-transparent nature of DNN, exploring the mechanism of adversarial attack and improving the adversarial robustness have become a significant theme of deep learning.

\begin{figure}
\centering
\includegraphics[height=6cm, width = 7.0cm]{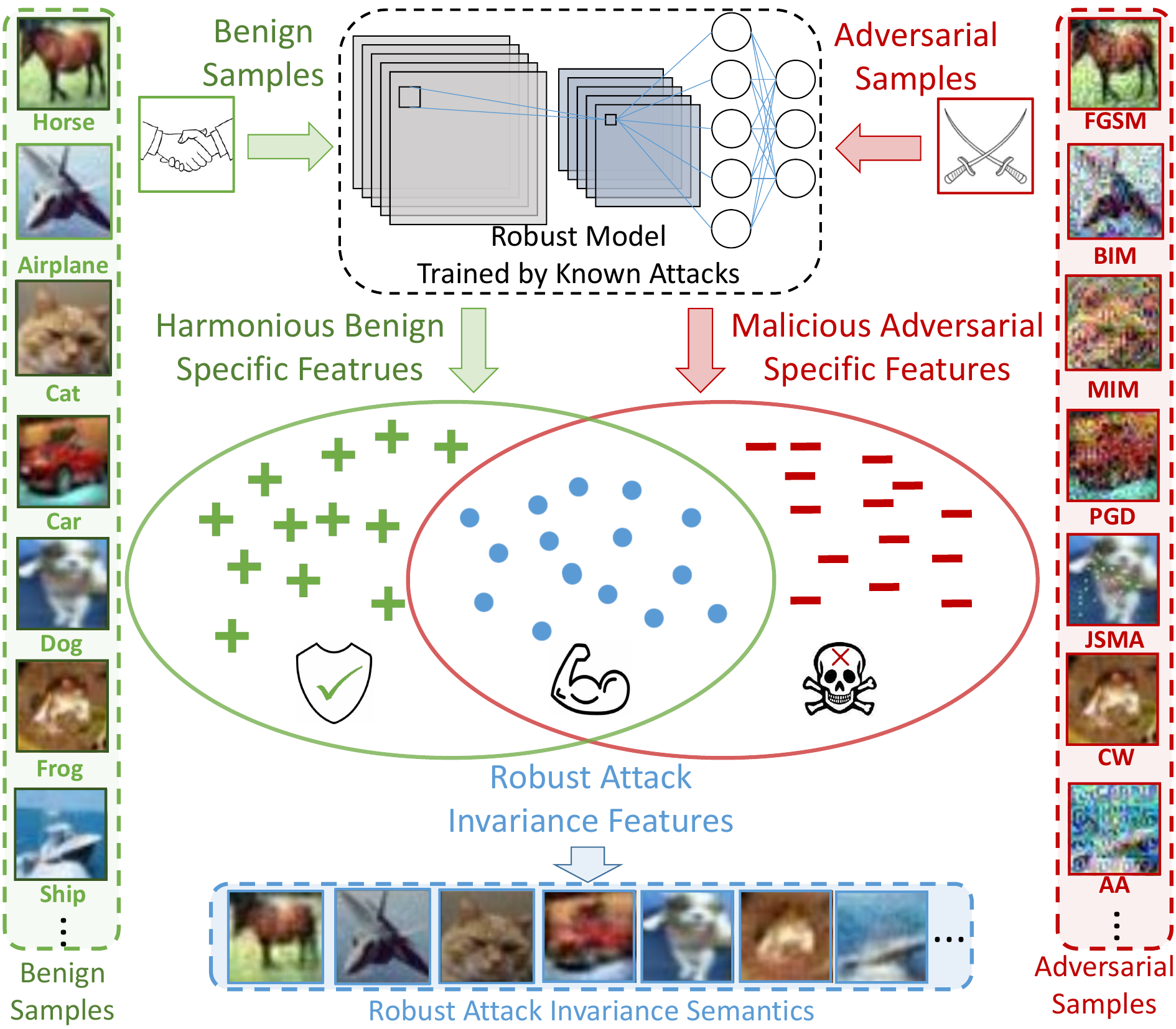}
\caption{An intuitive interpretation of attack invariant features. Humans are good at extracting robust invariant features from cats (such as beard and ears).  We argue that there exist such kind of Robust Attack Invariant Features (the blue points) between various adversarial attacks and benign samples, which only associate robust semantic rather than malicious perturbations. In this sense, the so-called benign and adversarial samples add harmonious benign features (the green +) and malicious adversarial features (the red -) to the invariant feature, respectively. 
}
\label{inv}
\end{figure}

\begin{figure*}
\centering
\vspace{-0.1cm}
\includegraphics[width = 14.5cm]{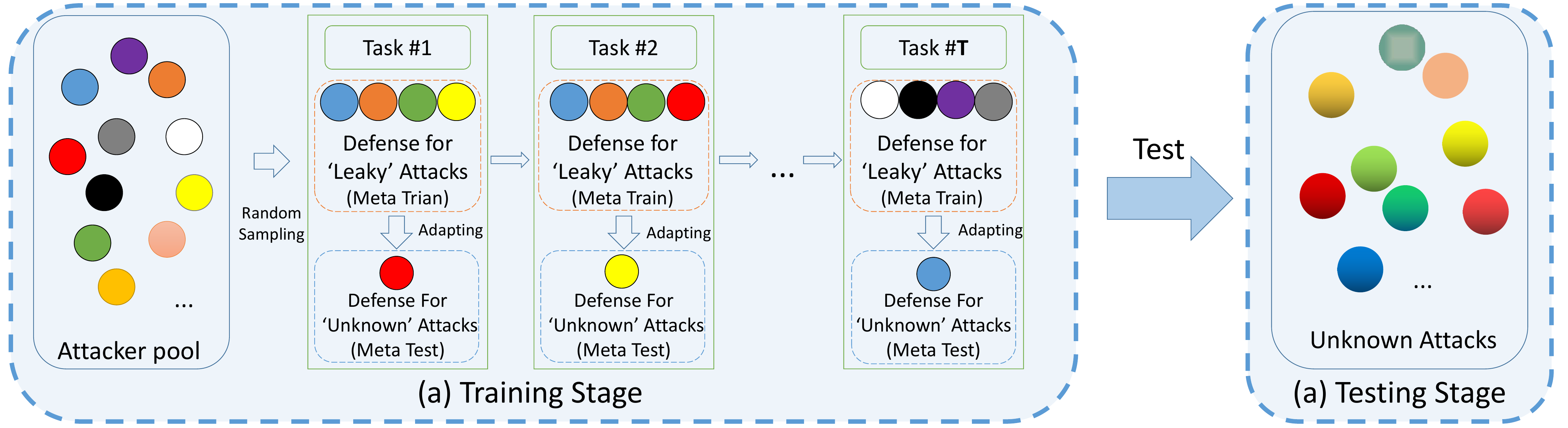}
\caption{Schematic diagram of MID's training process. For each epoch, MID can be divided into two stages: Meta-Train and Meta-Test. (1) In the Meta-Train phase, we randomly extract $n-1$ attack combinations from the \emph{Attacker Pool} as the simulated leaky attacks to train the robustness to known attacks. (2) During Meta-test phase, we choose the $n^{th}$ attack not used during Meta-Train as the simulated unknown attack to train the generalizable robustness to unknown attacks. The words with single quotation marks mean that they are simulated known/unknown attacks during the meta learning process, and the real unknown attack are never seen in the whole meta learning process.}
\label{fig:MID1}
\end{figure*}

Plenty of work are proposed to resist the impact of adversarial noise on DNN models \cite{goodfellow2014explaining,papernot2016distillation,liao2018defense,xie2017mitigating,zheng2016improving,shen2017ape,zhou2021towards}. Despite having various skills, the existing defense methods are basically facing \emph{Leaky Attacks}, i.e., the defender fully understands the attacker's attack strategy conducive to defense. However, over-reliance on known attacks during training makes the model defend only against those specific attacks, but overlooks various unknown attacks. In short, the traditional defense is not generalizable to unseen attacks, and DNN robustness is still a shy.
Unfortunately, this problem in real-world scenario seems to be ill-posed, because we neither know which models are robust nor which attacks we will encounter. Thus, to deal with the dilemma, we propose the concept of \textit{generalizable robustness}, towards the adversarial robustness to both known and unknown attacks. Notably, the known/unknown attacks depend on whether the attacker (e.g., attack strategy) being defended is available or not during training.

A subject in cognitive science studies the problem similar to our task, which tries to explain an interesting question: \textit{Why can human beings still accurately predict the identity of the face even with various, unknown expressions?} In short, this is because our brain is good at extracting expression-invariant identity features rather than fragile irrelevant features such as facial expression~\cite{mishkin1982contribution,kanwisher1997fusiform}. Similarly, after learning the prior knowledge about cats and tigers, one can easily infer that \emph{lynx} belongs to the \emph{cat} family, although we may not know its true name is \emph{lynx}. That is because humans are good at extracting invariant features in cats, such as whiskers, ears and facial contours.
This has been a fact that human beings can ignore various adversarial perturbations, but naturally focus on the meaningful and stably invariant semantic features. This enlightens us to explore attack-invariant features toward improving generalizable robustness of DNN to unknown attacks as indicated in Fig.\ref{inv}.

Toward the goal of domain invariance, domain generalization (DG) \cite{khosla2012undoing,muandet2013domain,ghifary2015domain} and meta-learning \cite{finn2017model,li2018learning} have made a progress. DG supposes that only the data from known source domain(s) can be fed into the model for training, which aims to obtain the generalization ability to unseen target domains. This means that the model needs to learn the domain-invariant features from the source domains, since such features are more stable and generalizable among various domains. Similarly, toward the goal of generalizable robustness, a robust model should be capable of learning attack-invariant features from several known attacks to obtain the generalization, since such kind of attack invariance shall also be stable and generalizable among various attacks. \cite{amich2022rethinking} even proposes that adversarial attack is a special form of the out-of-distribution (OOD) problem, and experimentally proves this conjecture via a domain classifier. Despite the lack of rigorous mathematical proofs, we give an intuition that there is some inevitable connection between OOD and adversarial attacks.

Meta-learning gives model the ability of \emph{learning to learn} through an adaptive confirmatory learning idea, which also fits our objective to learn a learning pattern adaptive to various attacks. In other words, meta learning not only tries to learn intrinsic knowledge, but to tackle the generalisation toward unknown tasks by simulating the verification task during training. In this sense,  meta learning acts like a regularization to reduce the structural risk, in order to find the most generalized function family in the hypothesis space.

Based on the above principles, we propose a new adversarial defense method named \textbf{Meta Invariance Defense (MID)}. In short, we combine the two-stage meta-learn mechanism with multiple consistency distillation constraints, and train a student encoder to extract attack-invariant features from the attacker pool by a teacher-student distillation paradigm. Finally, a robust defense framework against various attacks is formulated.

As shown in Fig.\ref{fig:MID1}, to select the stable framework, the training process of MID is divided into two stages: Defense for Simulated Leaky Attacks (Meta-Train) and Defense for Simulated Unknown Attacks (Meta-Test). Suppose we have an \emph{Attacker Pool} composed of multiple attack methods, and three basic steps are conducted in each training iteration:

\textbf{Step 1}: During Meta-Train, we randomly sample a set of attacks from the \emph{Attacker Pool}, and simulate the robust training against the (simulated) known attacks. At this stage, a temporary model is obtained by performing a single step gradient descent optimization on the original model, and the loss for meta train is obtained by evaluating the original model under (simulated) known attacks.

\textbf{Step 2}: During Meta-Test, we select the attack not used in Step 1 from the \emph{Attacker Pool}, then simulate the testing by the (simulated) unknown attacks for the temporary model. At this stage, the loss for meta test is obtained by evaluating the temporary model under (simulated) unknown attacks.

\textbf{Step 3}: We combine the losses of meta-train and meta-test, and train the original model with robustness generalizing both known and unknown attacks.

In this way, it may learn better generalization patterns by selecting the stable parameters having similar activations between known and unknown attacks. However, the framework of meta learning can only screen out stable parameters to all tasks (i.e., adversarial defense against various attacks), but unable to obtain attack-invariant features. Moreover, we cannot fine-tune the defense model for a specific task like \cite{finn2017model} to extract attack-specific features, due to the zero-shot nature of adversarial defense during deployment.
Thus, to further learn the attack-invariant features, we propose a multi-consistency distillation protocol to the meta-process, in order to distill the attack-invariant feature from the teacher network to the student network. On one hand, distillation mitigates the model-specific perturbation attack since it contains some effects of distillation defense~\cite{papernot2016distillation}. On the other hand, the multi-consistency protocol facilitates the semantic consistency with benign samples, which is exactly the attack-invariant feature among various attacks.

After the meta optimization, MID is tested on the real unknown attacks. Note that all the attacks used in the training process are regarded as known attacks, while the unknown attack is never seen during training.
 In summary, our contributions are three-fold:

(1) We tackle a twin challenge of \textit{robustness} and \textit{generalization} of adversarial defense and propose a Meta Invariance Defense (MID) approach towards generalizable robustness to unknown attacks. By simulating the defense against the simulated leaky attacks and unknown attacks in an iterative manner, parameters with similar activations for both known and unknown attacks are selected and retained.

(2) A multi-consistency distillation protocol containing adversarial, label and cyclic consistency constraints is proposed in MID model, which facilitates the learning and optimization of attack-invariant features.

(3) Theoretical and empirical analysis verify the feasibility and superiority of MID over SoTA adversarial robust approaches. More qualitative experiments also well interpret a set of conjectures, and our intuition is validated.

\section{Related Work}
\subsection{Adversarial Attack and Defense}
\subsubsection{Adversarial Attacks}
For CV tasks, adversarial attacks mean the human imperceptible perturbations, but could definitely mislead the DNN models and make incorrect prediction. According to whether the attacker can access the target model, adversarial attacks are generally divided into \emph{white-box} and \emph{black-box}.

\emph{White Box Attack} means the attacker can have all information of the target model, such as structure, parameters, gradient, etc. Several popular white box attacks include gradient based (e.g., FGSM \cite{goodfellow2014explaining}, BIM \cite{kurakin2016adversarial}, PGD \cite{madry2017towards}), classification layer perturbation based (e.g., DeepFool \cite{moosavi2016deepfool}) and conditional optimization based (e.g., CW \cite{carlini2017towards}, OnePixel \cite{su2019one}).

\emph{Black Box Attack} means the attacker has little prior knowledge about the model being attacked, and can be roughly divided into score-based, decision-making based, and transfer-based. The score-based attack assumes that the attacker can obtain the classification probabilities of each class of the target model (e.g., zoo \cite{chen2017zoo}). Decision-based black box attack assumes that the attacker can only obtain the one-hot prediction rather than scores (e.g., Boundary attacks \cite{brendel2017decision}). The transfer-based attack assumes that the attacker cannot access any information of the target model, and \cite{dong2018boosting,xie2019improving,dong2019evading,zou2020improving,lin2019nesterov} try to design a more transferable attack. Basically, the transfer-based black box attacks are often used to evaluate the adversarial robustness of DNN.

\subsubsection{Adversarial Defense}
In order to improve the robustness of DNN under adversarial attacks, the idea of defense in the community is quite divergent.
The empirical defense mainly includes adversarial training  \cite{goodfellow2014explaining, madry2017towards,kurakin2016adversarial} and defensive distillation \cite{zhang2019theoretically,wang2020improving,goldblum2020adversarially,zhu2021reliable}. Recently, \cite{jia2022adversarial} proposes a learnable attack strategy to enhance adversarial robustness.\cite{mustafa2020deeply} proposes to perturb the intermediate feature representations of deep networks to improve the adversarial robustness. \cite{wang2021agkd} proposes a defense method for bidirectional metric learning guided by an attention mechanism. \cite{pang2020bag} fully explores the potential of adversarial training using various tricks, and \cite{zi2021revisiting} proposes Robust Soft Label Adversarial Distillation to fully exploit the robust soft labels produced by a robust teacher model to guide the student model. Heuristic defense forms are more diverse. \cite{xie2017mitigating} proposes to randomly transform the inputs and eliminate specific adversarial pixels. Model Ensemble \cite{bagnall2017training} uses multiple models to vote and alleviate the model-specific attack.  \cite{zheng2016improving} proposes to shorten the coding distance between the original sample and the noisy sample. \cite{liao2018defense} attempts to introduce residual connection in low and high-dimensional feature layers. \cite{yang2021class} proposes that adversarial perturbations mainly exist in class-dependent information and designs adversarial sample detection for defense.  \cite{mao2021adversarial} utilizes the structural information to achieve defense in inference stage. Besides, theoretically-guaranteed defense remains an open problem, and \cite{raghunathan2018certified} proves that under the specific target model, the certified defense can ensure adversarial robustness.

It is noteworthy that most of the defense methods follow the assumption of \emph{Leaky Attack}, i.e., the defender fully understands the attack strategy. Adversarial training means the adversarial examples can be jointly used to train a robust model together with the benign samples. Obviously, the defense performance of such methods depends on the adversarial samples. Therefore, adversarial training based defense lacks flexibility and adaptability to unknown attacks.
In actual deployment scenario, the strategy of the attacker is completely unknown, which is insufficient to train the robustness of DNN to several known attacks. However, the generalization of robustness to unknown attacks is seriously overlooked. This is the key motivation of this paper.

\subsection{Meta Learning}As a novel learning to learn framework, the potential of meta learning in various fields has yet tapped. Meta learning tries to make the model obtain the ability of \emph{learning to learn}. It mainly focuses on how to use seen data to discover a learning pattern that also fits unseen data. When encountering a new task, the model only needs a few additional data to quickly adapt to the new task. \cite{hsu2018unsupervised} introduces the meta training strategy into unsupervised learning. \cite{schweighofer2003meta} proposes to combine Meta Learning with reinforcement learning. \cite{wang2019meta} uses meta learning to design a few-shot object detection framework. \cite{finn2017model} proposes to train a few-shot model that can quickly adapt to new tasks. \cite{li2018learning} proposes to learn domain-invariant features for domain generalization. \cite{yuan2021meta} learns the invariant information among different models to achieve a stronger black box attack that can transfer among models. In fact, these work can be considered to learn intrinsic invariance by exploiting meta learning scheme. For example, \cite{finn2017model} learns the invariance among different tasks, \cite{li2018learning} learns the invariance among different domains, and \cite{yuan2021meta} learns the invariance among different models. A natural idea is to learn the invariance among different attacks by exploring the meta learning framework in adversarial defense.

However, a fact is that meta learning can only screen robust parameters from the hypothesis space, but neglects the representation ability of specific tasks and the attack-invariance is still tricky. Therefore, in order to improve the attack-invariance representation, a multi-consistency distillation protocol is formulated in the meta training paradigm. The twin challenge of generalization and robustness to adversarial attacks is unified and tackled.

\section{Methodology}
\begin{figure*}
\centering
\includegraphics[width = 14.5cm]{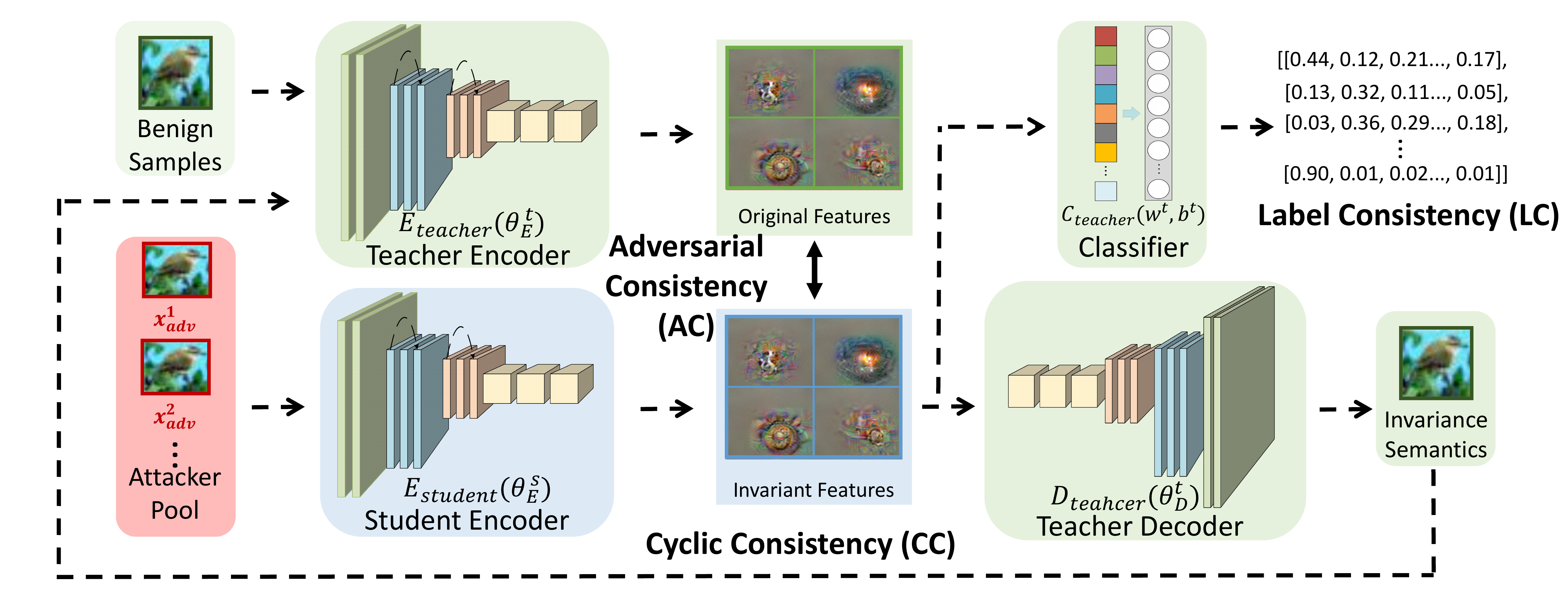}
\caption{The framework of Meta Invariance Defense. The teacher model with green shades is fixed in the training process, while the student module with blue shade is updated online based on the proposed multi-consistency distillation via the meta learning paradigm. The output of teacher decode is the regenerated robust sample. The final test is conducted by the robust student encoder $E_{student}$ and classifier $C_{teacher}(w,b)$.}
\label{fig:MID2}
\end{figure*}

\subsection{Preliminaries}

We use $E(\theta)$ to represent the feature extractor and $C(w, b)$ the classifier, while a complete model $F(\theta,w,b)$ is composed of an encoder $E(\theta)$ and a classifier $C(w,b)$. For an image recognition system, an input image is represented by high-dimensional feature through $E(\theta)$ and then predicted by the classifier $C(w,b)$. Let $x$ and $y$ respectively represent the benign samples and labels sampled from the joint distribution $P(X,Y)$, and each $x_{i}$ has a unique label $y_{i}$. The objective of an adversarial attacker can be expressed as a conditional optimization problem:
\begin{footnotesize}
\begin{equation}
\begin{aligned}
\underset{\mathrm{x}^{adv}}{\arg \max }  J\left(C\left  (E\left(x^{a d v},  \theta\right), w, b\right),y \right),
 \text { s.t. }\left\|x^{a d v}-x\right\|_{p} \leq \varepsilon
\end{aligned}
\label{eq1}
\end{equation}
\end{footnotesize}
where $J(\cdot)$ means the classification loss and $\varepsilon$ represents the upper bound of the perturbation under $l_{p}$-norm. The goal of the attacker is to fool the target model with visual imperceptible perturbation, such that the fooled model can make seriously wrong predictions. In this paper, we focus on the model defense against adversarial attacks and propose a simple but powerful adversarial defense framework for both the benign and attack-agnostic adversarial samples. The objective of defense is written as:
\begin{equation}
\begin{aligned}
C(E(x^{adv},\theta),w,b)=C(E(x,\theta),w,b)
\end{aligned}
\label{eq2}
\end{equation}
where $x^{adv}$ denotes attack-agnostic adversarial sample, which is only required to satisfy the condition in Eq.~\ref{eq1}.

\subsection{Meta Invariance Defense}
To achieve multi-consistency distillation, MID includes a student feature extractor (encoder) $E_{student}(\theta_{E}^{s})$ and a teacher network $F_{teacher }\left(\theta_{E}^{t}, w^{t}, b^{t}, \theta_{D}^{t}\right)$ (as shown in Fig.\ref{fig:MID2}). $F_{teacher}\left(\theta_{E}^{t}, w^{t}, b^{t}, \theta_{D}^{t}\right)$ includes a teacher encoder $E_{teacher}(\theta_{E}^{t})$, a teacher classifier $C_{teacher}(w,b)$ and a teacher decoder $D_{teacher}(\theta_{D})$. We use $F_{\text {teacher }}$ model to supervise the adversarial robustness training of $E_{student}(\theta_{E}^{s})$ based on the proposed multi-consistency distillation protocol.

Notice that the teacher model is expected to not only classification friendly, but also be capable to learn the manifold distribution of the inputs. Since the adversarial sample is considered to be an adjacent distribution on the input manifold, learning real data manifold can make DNN adapt to adversarial samples manifold distribution to a certain extent~\cite{meng2017magnet}. Auto-Encoder is an effective method to learn data manifold, so the $F_{\text {teacher }}\left(\theta_{E}^{t}, w^{t}, b^{t}, \theta_{D}^{t}\right)$ model is supervised by classification and reconstruction loss:
\begin{equation}
\begin{aligned}
L_{teacher} = L_{cls}^t(x,y,\theta_{E}^{t}, w,b) + L_{rec}^t(x,\theta_{E}^{T},\theta_{D}^{t})
\end{aligned}
\label{eq3}
\end{equation}
where the two losses are represented as follows.
\begin{equation}
\begin{aligned}
&L_{cls}=-\sum_{n=1}^{N} y_{n} \cdot \log \left(C_{teacher}\left(E_{teacher}, w, b\right)\right),
\end{aligned}
\label{eq4}
\end{equation}
\begin{equation}
\small
\begin{aligned}
&L_{rec}=\sum_{n=1}^{N}\left\|D_{teacher}\left(E_{teacher}\left(x_{n}, \theta_{E}^{t}\right), \theta_{D}^{t}\right)-x_{n}\right\|_{2}
\end{aligned}
\label{eq5}
\end{equation}

We further use $F_{\text {teacher }}$  to supervise the training of $E_{student}$ for exploring the attack-invariant features through meta learning, and $E_{student}$ is our expected adversarial robust encoder together with the classifier $C_{teacher}$. Notice that the teacher module is pre-trained and fixed during training of student model.

As we mentioned, the framework of meta learning can only filter robust parameter groups without effective encoding ability, and the adversarial defense cannot perform down-stream fine-tuning after training like few-shot tasks. Thus we directly learn attack-invariant features through multi-consistency distillation in the meta learning process rather than fine-tuning after meta learning as \cite{finn2017model} did. Therefore, besides the exploration of sensitive parameters to unknown attacks through two-step defense, we further propose a multi-consistency constraint protocol to supervise the attack-invariant feature encoding of $E_{student}$. The multi-consistency protocol includes Adversarial Consistency (AC), Cyclic Consistency (CC) and Label Consistency (LC), as shown in Fig.\ref{fig:MID2}. We will introduce them in \emph{meta train} and \emph{meta test} respectively.

\textbf{Meta-Train for Leaky Attacks.} Assuming that there are $n$ known attacks in the attacker pool that can be accessed by trainers. In a epoch of meta training, we randomly select $n-1$ attacks from the \emph{Attacker Pool} as the simulated known attacks, and meta train the model for the robustness for the simulated known attacks. We propose a compositional objective, i.e., the multi-consistency distilled from the teacher model, which are presented as follows:

\emph{(1) Adversarial Consistency (AC).} In the process of meta training, we only feed the benign samples into the $E_{teacher}$, but feed the corresponding adversarial samples randomly sampled from the \emph{Attacker Pool} into the $E_{student}$. In order to directly constrain $E_{student}$ to learn features similar to $E_{teacher}$, the AC loss is formularized as KL-divergence between the features of $E_{teacher}$ and $E_{student}$:

\begin{footnotesize}
\begin{equation}
\begin{aligned}
L_{AC}=\sum_{n=1}^{N} \sum_{i=1}^{I} p\left(E_{\text {teacher}}\left(x_{i}, \theta^{t}\right)\right) \cdot \log \left(\frac{p\left(E_{\text {teacher }}\left(x_{i}, \theta^{t}\right)\right)}{p\left(E_{\text {student }}\left(\tilde{x}_{i}^{n}, \theta^{t}\right)\right)}\right)
\end{aligned}
\label{eq6}
\end{equation}
\end{footnotesize}
where $p(\cdot)$ represents the probability distribution and $\tilde{x}_{i}^{n}$ represents the $n-th$ adversarial version of the $i-th$ original sample. We hope that the model can learn similar feature with its original sample for each adversarial sample.

\emph{(2) Cyclic Consistency (CC).} To further ensure that the features learned by $E_{student}$ only contains true semantics without the features of adversarial perturbations, the Cyclic Consistency is designed. We propose to exploit the teacher decoder $D_{teacher}(\theta_{D})$ to decode the output from $E_{student}$ into robust samples $x_{reg}$ (regeneration), and again feed the regenerated robust image into the $E_{teacher}$ for re-encoding. Then we use KL-divergence to constrain the distribution similarity between the features of the regenerated image and the original image. The CC loss is presented as:
\begin{equation}
\footnotesize
\begin{aligned}
L_{CC}=\sum_{n=1}^{N} \sum_{i=1}^{I} p\left(E_{\text {teacher}}(x_i,\theta^t)\right) \cdot \log \left(\frac{p\left(E_{\text {teacher}}\left(x_{i}, \theta^{t}\right)\right)}{p\left(E_{\text {teacher }}\left(x_{reg_{i}^{n}}, \theta^{t}\right)\right)}\right)
\end{aligned}
\label{eq7}
\end{equation}
where $x_{reg_{i}^{n}}$ represents the image regenerated from the adversarial sample by different attacks. Through a feedback structure, we expect that the regenerated image is robust enough in representation, rather than clean enough in pixel.

\emph{(3) Label Consistency (LC).} Additionally, we expect the student encoder $E_{student}$ and teacher classifier $C(w,b)$ can accurately predict any adversarial sample sampled from the \emph{Attacker Pool}. Therefore, we propose cross-entropy based LC loss to learn the distribution similarity between the classifier outputs and the ground-truth labels.
\begin{small}
\begin{equation}
\begin{aligned}
L_{LC}=-\sum_{n=1}^{N} \sum_{i}^{I} y_{n} \cdot \log \left(C\left(E_{\text {student}}\left(\tilde{x}_{i}^{n}, y_{n}, \theta^{s}\right), w, b\right)\right)
\end{aligned}
\label{eq8}
\end{equation}
\end{small}
Therefore, the training objective of the proposed multi-consistency in meta-train stage is formalized as:
\begin{equation}
\begin{aligned}
L_{meta\_train}=  \omega_{\text {AC}}  \cdot L_{\text {AC}} (E_{student}, X_n)&
\\+~  \omega_{\text {CC}} \! \cdot L_{\text {CC }} (E_{student}, X_n)
\\+ \ \omega_{\text {LC}} \cdot L_{\text {LC}}(E_{student}, X_n)
\end{aligned}
\label{eq9}
\end{equation}
where $X_n$ refers to the ensemble attacks used in the Meta Train. With the above hybrid loss of meta-train, we perform a single step gradient descent on the current student encoder to obtain a temporary model $E_{student}^{\prime}$, based on which the following meta-test is performed.

\textbf{Meta-Test for Unknown Attacks.} To obtain the adaptability and robustness to unknown attacks, we further simulate unknown attacks for defense in the meta test stage. Simply, in meta-test phase, we select the attack $X'_n$ not used in meta-train from the \emph{Attacker Pool} and feed into the one-step-optimized temporary model $E_{student}^{\prime}$. Then we use the same objective function Eq.\ref{eq9} to re-evaluate the temporary model $E_{student}^{\prime}$, and the meta-test loss is:
\begin{equation}
\begin{aligned}
L_{meta\_test} =  \omega_{\text {AC}}  \cdot L_{\text {AC}} (E_{student}^{\prime}, X'_n)&
\\+~  \omega_{\text {CC}} \! \cdot L_{\text {CC }} (E_{student}^{\prime}, X'_n)
\\+ \ \omega_{\text {LC}} \cdot L_{\text {LC}}(E_{student}^{\prime}, X'_n)
\end{aligned}
\label{eq10}
\end{equation}
where $E_{student}^{\prime}$ is the temporary model obtained by one-step gradient descent during the meta-train stage, and $X'_n$ is the simulated unknown attack not used in meta-train.

\textbf{Overall Optimization Objective.} Our objective is giving the student network robustness to known attacks and adaptive generalization to unknown attacks, therefore the meta-train and meta-test should be jointly optimized for the real student encoder. The meta-test process actually simulates the student model's response to unknown attacks. It is an automatic cross-validation and robust parameter-screening, then the generalization is achieved by $E_{student}$. Thus we combine the loss of meta-train stage with that of meta-test stage to train the robust student encoder $E_{student}$. Thus, the final optimization goal for $E_{student}$ becomes:
\begin{equation}
\begin{aligned}
L_{all}= L_{meta\_train} + L_{meta\_test}
\end{aligned}
\label{eq11}
\end{equation}
In this way, the $E_{student}$ will find the non-robust parameters sensitive to both the known and unknown attacks, and learn attack-invariant features with multi-consistency constraints in an iterative manner. The detailed implementation of the proposed MID is summarized in \textbf{Algorithm 1}.

\begin{algorithm*}[tb]
\caption{Meta Invariance Defense}
\label{alg:algorithm}
\textbf{Require}: $Attacker Pool$ $P(\mathcal{A})$ including $\mathcal{N}$ attacks: $\mathcal{A}_0$, $\mathcal{A}_1$... $\mathcal{A}_{\mathcal{N}}$, and a fixed teacher model $F_{teacher}$ including teacher encoder $E_{teacher}{(\theta_{E}^{t})}$, teacher decoder $D_{teacher}(\theta_{D}^t)$ and classifier $C_{teacher}(w^t,b^t)$. \\
\textbf{Input}: Benign samples $x$, labels $y$, $F_{\text {teacher }}\left(\theta_{E}^{t}, w^{t}, b^{t}, \theta_{D}^{t}\right)$, and an initialized student encoder $E_{student}(\theta^s_E)$. \\
\textbf{Parameter}: Fixed teacher model parameters: $\theta_{E}^{t}, w^{t}, b^{t}, \theta_{D}^{t}$, and unfixed student model parameters: $\theta^s_E$.   \\
\textbf{Output}: Robust student encoder $E_{student}(\theta^s_E)$.
\begin{algorithmic}[1] 
\STATE Initializes the $\theta^s_E$ of $E_{student}(\theta^s_E)$.
\WHILE{not converge}
\STATE Randomly sample $\mathcal{N}$-1 attacks from the $Attacker Pool$ and formalize a new assemble attack $X_n$.
\FOR{all $X_n$}
\STATE Calculate the total meta-train loss $L^{X_n}_{meta\_train}(E_{student}(\theta^s_E))$ for each attack according to Eq.9:\\

$L_{meta\_train}(E_{student}(\theta^s_E)) = \sum_{{n=1}}^{\mathcal{N}-1} L^{X_n}_{meta\_train}(E_{student}(\theta^s_E), X_n)$.

\STATE Perform a single step gradient descent for meta updating and get a temporary model $E_{student}^{\prime}({\theta^s_E}^{\prime})$: \\ ${\theta^s_E}^{\prime}=\theta^s_E-\alpha \nabla_{\theta^s_E} \mathcal{L}_{meta\_train}(E_{student}(\theta^s_E),X_n)$.
\ENDFOR
\STATE Select the unused attack $X_n^{\prime}$ for the simulated optimization of unknown attack in meta-test stage, and combine the losses of meta-train and meta-test to jointly optimize the student encoder:\\
$\theta^s_E = \theta^s_E-\gamma \cdot \frac {\partial\left(\mathcal{L}_{meta\_train}(E_{student}(\theta^s_E),X_n)+\beta \mathcal{L}_{meta\_test} (E_{student}({\theta^s_E}^{\prime}),X_n^{\prime}) \right)}{\partial \theta^s_E}$\\

\ENDWHILE
\end{algorithmic}
\end{algorithm*}

\textbf{Model Test.} The MID model trained in Algorithm 1 will be tested on real unknown attacks, and the final robust model consists of the learned robust student encoder $E_{student}(\theta_{E}^{s})$ and the classifier $C_{teacher}(w, b)$.  At this point, we consider all the attacks used during training as known attacks including simulated known attacks in meta-train and simulated unknown attacks in meta-test. The real unknown attacks for test are never seen during training.

\textbf{Further Notes.} To be clear on the implementation, we further clarify that the meta-train and meta-test are two stages of MID training and conducted on the training set. The test set of unknown attacks for model test is completely unseen. The simulated unknown attack in the meta-test stage is obtained from the training set through a stochastic attack, which is feasible only if such attack is not used in the meta-train stage. Also, our goal of MID contains not only the adversarial robustness to known attacks, but also the generalization to unknown attacks.

\subsection{Theoretical Feasibility Analysis of MID}\label{feasibility}

In this section, we theoretically analyze why MID is effective from different perspectives.

\subsubsection{Perspective of Taylor Expansion} The objective function of MID can be summarized as:
\begin{small}
\begin{equation}
\begin{aligned}
 \min_{\theta} J\left(x^{\prime}, y, \theta\right)+J\left(x^{\prime \prime}, y, \theta-\alpha \cdot \nabla_{\theta} J\left(x^{\prime}, y, \theta\right)\right)
\end{aligned}
\label{eq12}
\end{equation}
\end{small}
where $x'$ means adversarial sample in meta-train, $x''$ is the simulated unknown attack, and $J$ is the multi-consistency loss in Eq.\ref{eq9}.  We further transform the second term in Eq.~\ref{eq12} by first-order Taylor expansion as follows:
\begin{small}
\begin{equation}
\begin{aligned}
\min_{\theta} J\left(x^{\prime \prime}, y, \theta\right)+\nabla_{\theta} J\left(x^{\prime \prime}, y, \theta\right) \cdot \left(-\alpha \nabla_{\theta} J\left(x^{\prime}, y, \theta\right)\right) \\ + o\left(-\alpha \nabla_{\theta} J\left(x^{\prime}, y, \theta\right)\right)
\end{aligned}
\label{eq13}
\end{equation}
\end{small}
and the total objective Eq. \ref{eq12} becomes:

\begin{equation}
\begin{aligned}
 \min_{\theta} &J\left(x^{\prime}, y, \theta\right)  + J\left(x^{\prime \prime}, y, \theta\right)- \\ & \alpha \cdot \nabla_{\theta} J\left(x^{\prime \prime}, y, \theta\right) \cdot\nabla_{\theta} J\left(x^{\prime}, y, \theta\right) + o\left(-\alpha \nabla_{\theta} J\left(x^{\prime}, y, \theta\right)\right)
\end{aligned}
\label{eq14}
\end{equation}

This reveals the attack-invariant essence of our objective: (1) It minimizes the loss of known attack (the $1^{st}$ item in Eq.\ref{eq14}) in meta-train and unknown attack (the $2^{nd}$ item in Eq.\ref{eq14}) in meta-test. (2) The $3^{rd}$ term is amount to maximizing the cosine similarity of the gradient between known and unknown attacks. In other words, it can learn the gradient commonality between known and unknown attacks towards learning attack-invariance. However, directly using Eq. \ref{eq14} may be insufficient since it not only ignores the smoothness brought by higher-order derivative optimization, but also overlooks the implicit parameter regularization effect based on the adaptive training pattern (see Sec. \ref{sec4_3}). This equals to only constraining the gradient distances for several kinds of attacks for generalization like ERM \cite{vapnik1999overview}. We thus propose to train the full loss function in Eq. \ref{eq12}.

Traditional defense methods can only access known attacks but do not have generalization ability to new attacks.
In MID, maximizing gradient similarity means that we constrain the model optimization towards a common direction for both known and unknown attacks, and suppress the parameters that are difficult to uniformly optimize for all attacks (i.e., attack-specific parameters). Then, supervised by the additional consistency constraints, the attack-agnostic parameters clearly reveal the attack-invariant features. Thus, MID not only favors the known attacks, but discover invariant information among various attacks, so the generalizable robustness for unknown attacks is achieved.

\subsubsection{Perspective of Manifold Interpretation}
Manifold learning \cite{tenenbaum2000global,belkin2001laplacian,roweis2000nonlinear} attempts to recover the low dimensional manifold structure from the high-dimensional space and describe the spatial neighborhood relationship between each sample, so as to realize dimension reduction or data visualization. Traditional manifold learning is mostly based on exquisite manual design, but the emergence of deep model provides a simple and effective preprocessing for manifold learning. For example, an AutoEncoder aims at learning the distributed representation and nearest neighbor graph from the original high dimensional space \cite{goodfellow2016deep}. An original image of 224*224*3 is represented as a 512-dimensional feature representation by the encoder. After effective regularization, one dimension or a combination of several dimensions can even represent some significant semantics, such as edge, texture, and even the shape of a vehicle's tire. Therefore, by modifying some semantics in the feature maps, a decoded image from the decoder can be exactly the adjacent distribution of the original image (e.g., put glasses on a person's face, or make a smiling face expressionless). Compared with the original image, the low dimensional feature manifold is obviously easier to capture through a huge amount of data~\cite{scholkopf1998nonlinear,hinton2002stochastic,weinberger2006unsupervised}.

\begin{figure}
\centering
\includegraphics[height=2.5cm]{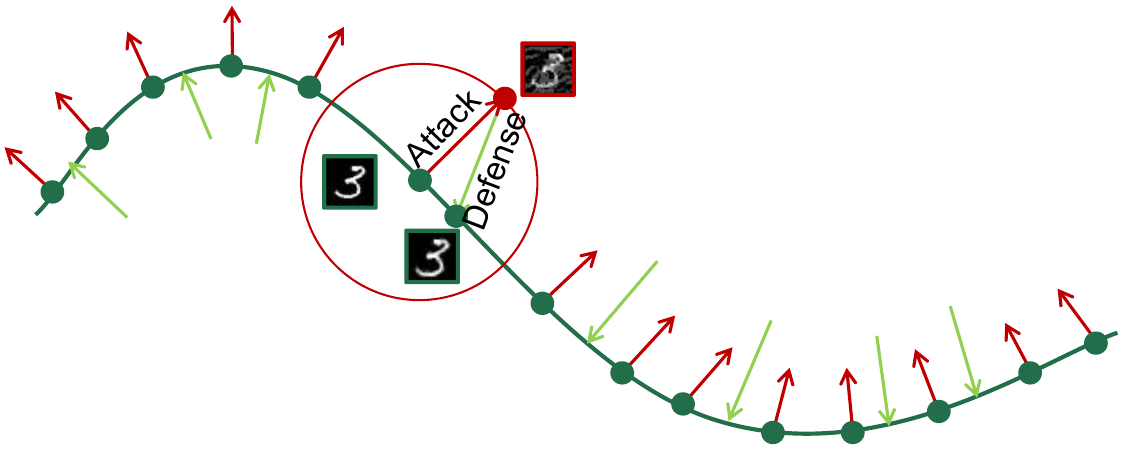}
\caption{Schematic diagram of manifold interpretation for MID. Gradient based adversarial samples push the benign samples away from the original manifold along the normal vector, since the gradient corresponds to the direction of the normal vector. MID aims to learn the similarity between the original features (benign samples) and neighbor features (i.e., adversarial samples) under a multi-consistency distillation via meta learning, and pulls the adversarial samples back to the original manifold.}
\label{fig:manifold}
\end{figure}

Further, the similarity metric (KL divergence) and classification loss (cross entropy) also describe the manifold of data. The smaller the entropy is, the tighter the manifold is. Thus it can be considered that the gradient of entropy corresponds to the normal vector of manifold curve. The gradient-based adversarial sample pushes the original sample away from the original distribution in the fastest direction (gradient direction can be considered as the normal vector), and the MID model attempts to transport the feature distribution of adversarial sample to the original manifold (as shown in Fig.\ref{fig:manifold}). On the one hand, the direct similarity constraint ($L_{AC}$) at the feature level ensures that the student model only learns the adversarial invariant features similar to the original features. On the other hand, the teacher model of MID contains an AutoEncoder module. The decoder decodes the features learned by the student model and further facilitates the student network to learn the neighbour distribution of the original samples ($L_{CC}$), i.e., robust sample regeneration. The qualitative analysis in the experiment section verifies our conjecture by decoding the features of the student encoder and t-SNE visualization.

\subsubsection{Perspective of High-order Optimization}
To fit the target task, the loss of a high-performance model is expected to be close to 0 at the optimal point. To further avoid the ill-condition problems, we hope the first-derivative (Jacobian matrix) of a robust model at the optimal point is also close to \textbf{0} (null matrix), which means we want the model to be smooth at the optimal point, since a smoother model has more generalizable robustness. To interpret this perspective, we pose three questions.

\begin{figure}
\centering
\includegraphics[height=2.8cm,width=9cm]{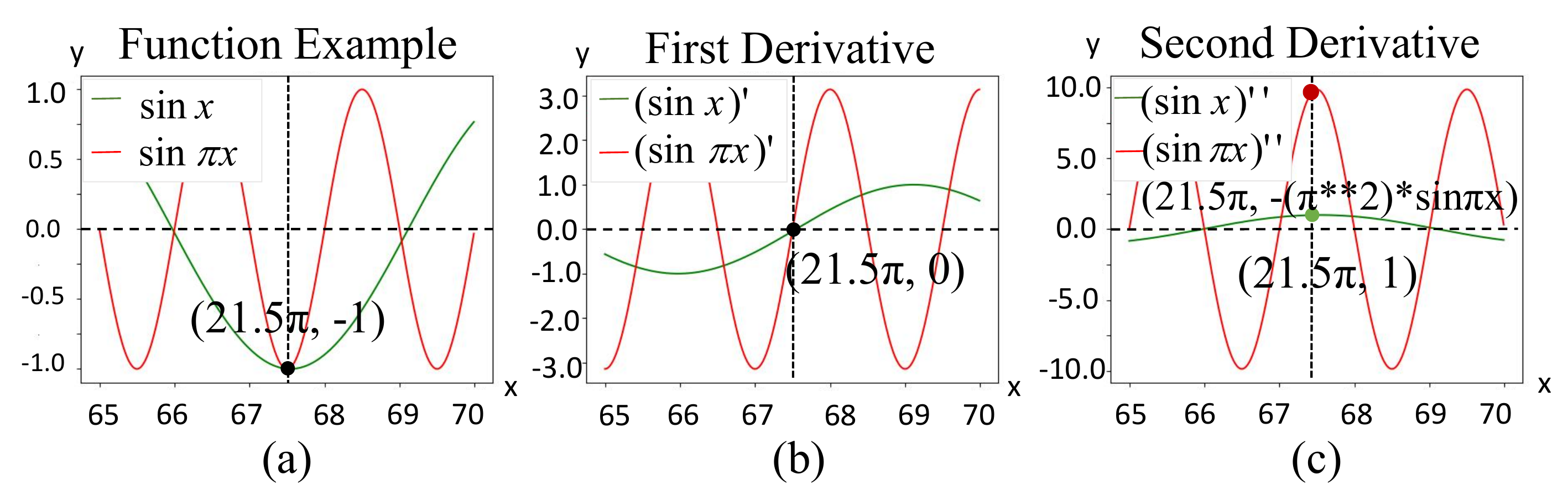}
\vspace{-0.4cm}
\caption{A toy example between the second derivative (Hessian matrix) and the robustness of the function. We take $sin(x)$ and $sin(\pi x)$ as an example. 
At the minimum point, $sin(x)$ and $sin(\pi x)$  have the same minimum value and first-derivative value, but $sin(x)$ is more robust than $sin(\pi x)$ since $sin(x)$ has a smaller second-derivative value. Thus both the curve of $sin(x)$ and its first derivative are smoother. MID realizes implicit regularization to the second-derivative of loss function.}

\label{fig:flat}
\end{figure}

\textbf{Why can a flat minimum point provide generalization?} Cha et al.~\cite{cha2021swad} provide a theoretical illustration, which demonstrates that models with flatter minimal points can generalize well to the out-of-distribution (OOD) samples. We also give an intuitive explanation in Fig.\ref{fig:flat} (a): when the inputs slightly shift, the flatter function has less shift in its output and  shows better stability and generalization to the slightly shifted input sample.

\textbf{Why does higher-order optimization enable models to find flatter minima?} On the one hand, since vanilla models are often optimized with first derivatives, while first-derivatives are often optimized by second-derivatives, so second-order optimization can help find smaller first-derivatives where the model is more stable to the shifted inputs. This results in better robustness and generalization~\cite{ross2018improving}. On the other hand, for a differentiable model, the first-derivative near the saddle point (the suspected minimal point) is often close to 0. A smaller second-derivative means the first-derivative varies smoothly and slowly, which means the model is flatter at this saddle point. For the $g_1(x)=sin(x)$ and $g_2(x)=sin(\pi x)$ in Fig.\ref{fig:flat}, despite the same saddle points, $sin(x)$ has smaller second-derivatives and hence smoother first-derivatives, which shows more stable saddle points. Thus, higher-order optimization can bring flatter minima.

\textbf{Why can MID introduce higher-order optimization?} On the one hand, the Taylor expansion in Eq. \ref{eq13} illustrates the existence of higher-order terms in MID. On the other hand, similar to the vanilla loss optimized by the first-order gradient, the first-order gradient of MID loss is optimized by the second-order gradient. Specifically, in meta-train stage of MID, the model needs to calculate the first-order gradient (Jacobian matrix) after performing one-step gradient descent, and performs another step gradient descent in the meta-test stage, which is analogy to an implicit second-order derivative optimization, formalized as:
\begin{equation}
\footnotesize
\begin{aligned}
&\theta^{\prime \prime}=\theta-\alpha \cdot \nabla_{\theta} J\left(x^{\prime}, y, \theta\right)-\beta \cdot \nabla_{\theta^{\prime}} J\left(x^{\prime \prime}, y, \theta-\alpha \cdot \nabla_{\theta} J\left(x^{\prime}, y, \theta\right)\right) \\
&\approx \theta-\nabla_{\theta}\left(\alpha \cdot \partial J\left(x^{\prime}, y, \theta\right)+\beta \cdot \partial J\left(x^{\prime \prime}, y, \theta-\alpha \cdot \nabla_{\theta} J\left(x^{\prime}, y, \theta\right)\right)\right.
\end{aligned}
\label{eq15}
\end{equation}

Note that Eq.\ref{eq15} is just an approximation. This explains why our method can improve the robustness of both known and unknown attacks simultaneously. By calculating the Hessian matrix of MID loss, we induce the model to find a flat minima where the Jacobian matrix is not only close to zero but also flat, which enables MID to be clearly robust towards adversarial perturbations.

\section{Experiments}

\begin{figure*}
\centering
\includegraphics[height=2.5cm,width=16.7cm]{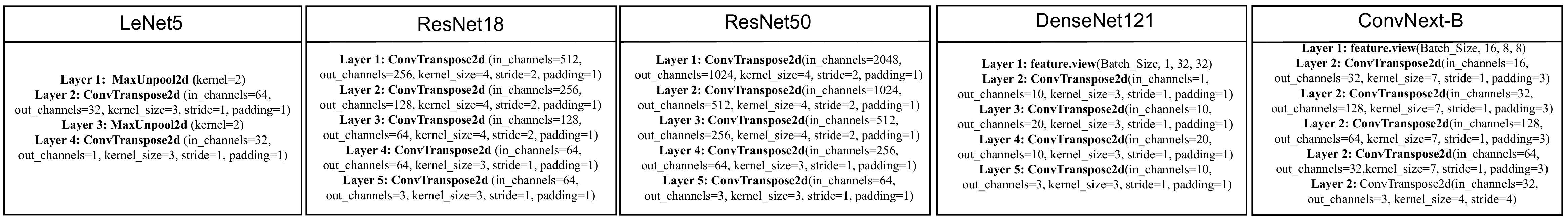}
\vspace{-0.2cm}
\caption{Different upsampling layers are designed for each backbone. For the relatively simple LeNet5, we adopt a completely symmetrical structure. For other complex backbones, the decoder is obtained by calculating the output of the de-convolution layer. The BN layer and ReLU are also adopted after each non-output upsampling layer (except LeNet5), while Tangent (Tanh) function is used as the activate function for the output layer.}
\label{fig:dec}
\end{figure*}
\subsection{Experimental Setup}
\textbf{Datasets.} Eight commonly used datasets for evaluation. MNIST has 10 classes of handwritten digit images. FashionMNIST is composed of the scanned pictures from 10 common clothes. Both MNIST and FashionMNIST are composed of single channel black-and-white images and the image size is 1x28x28. CIFAR-10 has 10 classes of common objects images containing 50,000 training images and 10,000 test images. The CIFAR-100 dataset has 100 classes. Each class has 600 color images, of which 500 are used as the train set and 100 as the test set. SVHN is a dataset composed of Street View of House Number, where 73257 pictures are used for training, 26032 pictures are used for testing, and 531131 additional and slightly less difficult samples are used as additional training data. Notably, CIFAR10, CIFAR100 and SVHN are composed of three channel color images, and the image size is 3x32x32. Besides, we also conduct on 3 large-scale datasets: TinyImageNet-200, ImageNet-100 and ImageNet-1K. TinyImageNet-200 has 200 classes, where each class has 500 training images, 50 validation images and 50 test images. ImageNet-100 is a subset of ImageNet-1K, where each category contains over 500 images in size of 3x224x224. ImageNet-1K is a commonly used large dataset containing over 1,200,000 images in size of 3x224x224.

\textbf{Backbones and Baselines. }We consider different backbone and baseline models for each dataset:

$\bullet$ MNIST: we use LeNet5 \cite{lecun1998gradient} as the target network of MNIST for digital recognition task.

$\bullet$ FashionMNIST: we also use LeNet5 \cite{lecun1998gradient} as the target network for FashionMNIST since FashionMNIST is similar to MNIST which also consists of single channel images.

$\bullet$ CIFAR10: we use ResNet18 \cite{he2016deep} as the target network of CIFAR10 for the multi-channel image classification task.

$\bullet$ SVHN:  we use ResNet50 \cite{he2016deep} as the target network of SVHN since it is a more challenging task than CIFAR10.

$\bullet$ CIFAR100: we use DenseNet121 \cite{huang2017densely} as the target network of CIFAR100 since the labels are much richer, but the intra-class samples are sparser.

$\bullet$ Tiny-ImageNet: we use ConvNext-B \cite{liu2022convnet} as the target network of Tiny-ImageNet.

$\bullet$ ImageNet100: we use Visual Transformer (small) \cite{dosovitskiy2020image} as the target network of ImageNet100. We only use adversarial consistency and label consistency during distillation.

$\bullet$ ImageNet-1K: we use Visual Transformer (base) \cite{dosovitskiy2020image} as the target network. We only use adversarial consistency and label consistency and we conduct the popular `pre-training $\&$ fine-tuning' training paradigm for comparison.

\begin{figure}
\centering
\includegraphics[height=2.2cm, width=7.5cm]{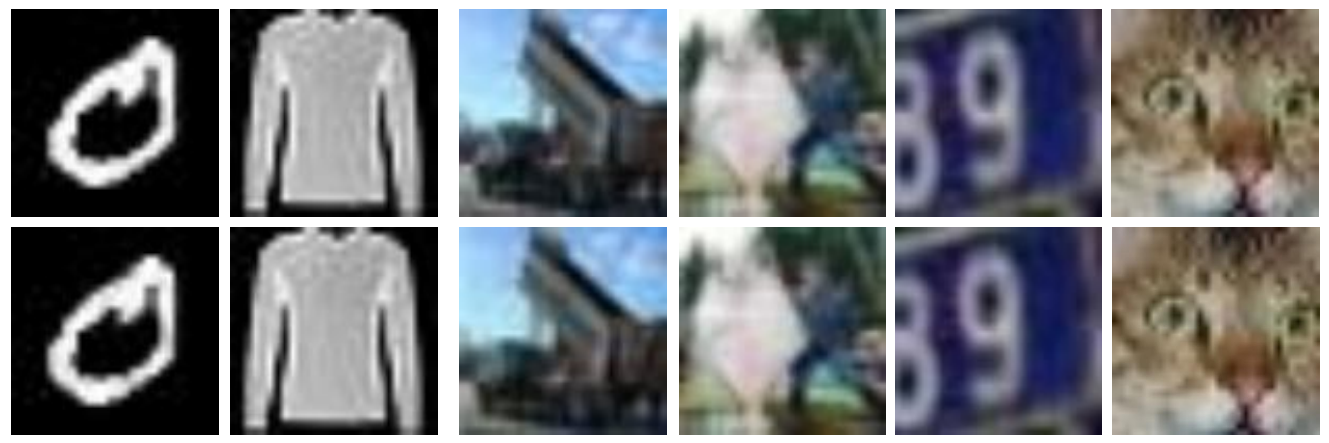}
\caption{Decoded samples from the teacher model.  The $1^{st}$ row shows some raw examples from the first six datasets and the $2^{nd}$ row consists of the reconstructed benign samples decoded by the teacher decoder. 
}
\label{fig:teach}
\vspace{-0.2cm}
\end{figure}

\textbf{Teacher Models. }To learn discriminative representations, teacher encoders adopt the same backbones as the student encoder and are trained by the cross entropy classification loss and $L_1$-norm based reconstruction loss (except models for ImageNet). We follow the settings in \cite{papernot2016distillation, goldblum2020adversarially} and adopt identical teacher and student encoder structure, considering that MID actually learns attack invariant information by aligning the representations of the teacher and the student encoder, rather than transfer teacher knowledge or answers to the student. The performance of larger teacher models than student model is discussed later.
The final teacher module $F_{\text {teacher }}\left(\theta_{E}^{t}, w^{t}, b^{t}, \theta_{D}^{t}\right)$ consists of a teacher encoder $E_{teacher}(\theta_{E}^{t})$, a classifier $C(w,b)$ and a decoder $D(\theta_{D})$. We design simple but effective decoders for each different backbones, as shown in the Fig.\ref{fig:dec}.
The classification ability of the teacher model is given in Tab.\ref{tab:teachercls}, and the decoding ability of the teacher decoder can be verified by Fig.\ref{fig:teach}. Therefore, on the one hand, the teacher models have qualified classification accuracy and can effectively distill the recognition ability to the student models. On the other hand, the learned features from the teacher encoder can be effectively restored by the decoder.
Therefore, the student encoder is thought to be adversarial robust and attack invariant, if the feature learned by the student encoder can also be decoded by the teacher decoder, since the output of the student encoder w.r.t an adversarial sample $x'$ is constrained to be similar to the teacher encoder w.r.t. its original sample $x$ via the adversarial consistency. Notice that all teacher modules are frozen and only the student modules are updated online.
\begin{table}[]
\footnotesize
\centering
\caption{The teacher models can achieve the same performance as the baseline where TI, I-100, and I-1k  represent Tiny-ImageNet, ImageNet-100, and ImageNet-1k respectively.}
\label{tab:teachercls}
\setlength{\tabcolsep}{0.1mm}{
\begin{tabular}{c|cccccccc}
\hline
Datasets                                        & MNIST   & \begin{tabular}[c]{@{}c@{}}Fashion\\ MNIST\end{tabular} & CIFAR10 & SVHN    & CIFAR100  & TI
&I-100 &I-1k \\ \hline
Baseline                                        & 99.16\% & 91.81\%                                                 & 83.99\% & 94.90\% & 60.36\%  & 53.70\% & 60.38\% & 81.91\%\\
\begin{tabular}[c]{@{}c@{}}Teacher\end{tabular} & 99.05\% & 92.02\%                                                 & 82.89\% & 94.04\% & 60.58\%  & 51.56\% & 60.38\% & 81.91\%\\ \hline
\end{tabular}}
\vspace{-0.2cm}
\end{table}

\begin{table*}[]
\caption{Defense for commonly evaluated white box attacks. Higher accuracy means better robustness. The best robustness is marked in bold.}
\label{tab:1}
\vspace{-0.2cm}
\scriptsize
\centering
\setlength{\tabcolsep}{1.4mm}{
\begin{tabular}{c|c|c|cccc|cccccccc|c}
  \hline
   &
   &
   &
  \multicolumn{4}{c|}{Known Attack} &
  \multicolumn{8}{c|}{Unknown Attack} \\ \cline{4-15}

\multirow{-2}{*}{} &
  \multirow{-2}{*}{Defense} &
  \multirow{-2}{*}{Benign} &
  $\rm PGD_N$ &
  $\rm PGD_T$ &
  $\rm MIM_N$ &
  $\rm MIM_T$ &

  $\rm FGSM_N$ &
  $\rm FGSM_T$ &
  $\rm BIM_N$ &
  $\rm BIM_T$ &
  $\rm CW_N$ &
  $\rm CW_T$ &
  $\rm AA_N$ &
  $\rm JSMA_T$ &
  \multirow{-2}{*}{Avg.}
  \\ \hline
 &
  None &
  \textbf{99.16} &
  0.61 &
  9.80 &
  0.67 &
  9.80 &
  19.30 &
  15.95 &
  0.67 &
  69.49 &
  0.66 &
  0.66 &
  0.00 &
  9.80 &
  13.19\\
 &
  $\rm AT'$\cite{madry2017towards} &
  98.70 &
  96.41 &
  98.40 &
  95.80 &
  97.26 &
  92.15 &
  96.08 &
  95.99 &
  86.98 &
  98.09 &
  97.98 &
  95.83 &
  52.57 &
  92.48\\
 &
  DST\cite{papernot2016distillation} &
  98.93 &
  0.65 &
  55.95 &
  18.71 &
  19.47 &
  32.46 &
  23.59 &
  45.69 &
  89.49 &
  91.22 &
  91.64 &
  0.25 &
  63.60 &
  48.58\\
 &
  DOA\cite{wu2019defending} &
  91.73 &
  34.77 &
  61.16 &
  69.18 &
  71.56 &
  80.16 &
  82.71 &
  83.97 &
  82.19 &
  88.14 &
  89.69 &
  7.14 &
  71.64 &
  70.31 \\
 &
  HGD\cite{liao2018defense} &
  90.4 &
  98.01 &
  97.85 &
  85.64 &
  88.78 &
  93.15 &
  93.97 &
  94.51 &
  93.99 &
  97.64 &
  96.54 &
  93.72 &
  45.89 &
  90.00 \\
 &
  RT$^{*}$\cite{xie2017mitigating} &
  84.57 &
  12.92 &
  39.53 &
  3.35 &
  19.76 &
  38.89 &
  28.86 &
  12.05 &
  88.30 &
  67.12 &
  62.4 &
  21.21 &
  48.69 &
  40.58 \\
 &
  ST\cite{zheng2016improving} &
  99.21 &
  21.97 &
  91.57 &
  32.48 &
  35.69 &
  35.22 &
  34.85 &
  1.19 &
  89.31 &
  62.18 &
  67.67 &
  10.70 &
  67.71  &
  49.98\\
 &
  APE\cite{shen2017ape} &
  99.05 &
  94.43 &
  98.71 &
  90.78 &
  97.16 &
  87.62 &
  81.16 &
  92.55 &
  89.49 &
  96.77 &
  96.68 &
  92.87 &
  49.40  &
  89.78\\
 &
  ARN\cite{zhou2021towards} &
  91.79 &
  98.15 &
  98.71 &
  92.56 &
  93.17 &
  93.12 &
  96.01 &
  95.37 &
  95.20 &
  98.25 &
  98.61 &
  97.62 &
  \textbf{83.25} &
  94.75 \\
\multirow{-10}{*}{\begin{tabular}[c]  {@{}c@{}}MNIST   \end{tabular}} &
  \textbf{MID} &
  98.83 &
  \textbf{99.01} &
  \textbf{99.20} &
  \textbf{99.05} &
  \textbf{98.15} &
  \textbf{93.47} &
  \textbf{96.72} &
  \textbf{98.90} &
  \textbf{98.78} &
  \textbf{98.30} &
  \textbf{98.70} &
  \textbf{98.24} &
  58.76 &
  \textbf{94.77} \\ \hline
 &
  None &
  \textbf{91.81} &
  4.04 &
  10.00 &
  6.79 &
  10.00 &
  14.77 &
  10.94 &
  6.63 &
  10.00 &
  5.94 &
  10.00 &
  0.00 &
  10.00 &
  14.68 \\
 &
  $\rm AT'$\cite{madry2017towards} &
  91.62 &
  78.80 &
  68.09 &
  91.49 &
  64.27 &
  56.92 &
  67.39 &
  61.72 &
  65.66 &
  89.18 &
  89.34 &
  92.87 &
  56.53 &
  74.91 \\
 &
  HGD\cite{liao2018defense} &
  81.07 &
  89.09 &
  87.13 &
  87.07 &
  79.55 &
  75.04 &
  84.13 &
  86.28 &
  84.39 &
  81.37 &
  84.98 &
  87.61 &
  43.78 &
  80.88 \\
 &
  APE\cite{shen2017ape} &
  89.07 &
  84.49 &
  84.36 &
  90.10 &
  87.23 &
  77.80 &
  84.59 &
  \textbf{92.55} &
  85.99 &
  86.62 &
  86.76 &
  \textbf{93.07} &
  50.02  &
  84.04\\
 &
  ARN\cite{zhou2021towards} &
  90.36 &
  88.89 &
  84.62 &
  89.36 &
  88.12 &
  69.93 &
  72.38 &
  86.72 &
  85.93 &
  86.97 &
  88.61 &
  91.51 &
  \textbf{62.15} &
  83.50 \\
\multirow{-6}{*}{\begin{tabular}[c]{@{}c@{}}Fashion \\ MNIST\end{tabular}} &
  \textbf{MID} &
  89.75 &
  \textbf{91.60} &
  \textbf{86.99} &
  \textbf{91.73} &
  \textbf{87.97} &
  \textbf{79.81} &
  \textbf{89.34} &
  91.69 &
  \textbf{86.89} &
  \textbf{89.62} &
  \textbf{89.68} &
  88.63 &
  52.86 &
  \textbf{85.56} \\ \hline
 &
  None &
  \textbf{83.99} &
  2.02 &
  10.23 &
  10.71 &
  10.04 &
  9.71 &
  7.44 &
  9.76 &
  54.08 &
  9.78 &
  0.00 &
  0.00 &
  10.00 &
  16.75 \\
 &
  $\rm AT'$\cite{madry2017towards} &
  63.24 &
  57.70 &
  53.22 &
  57.45 &
  46.28 &
  35.72 &
  35.77 &
  70.77 &
  51.58 &
  63.18 &
  62.69 &
  50.88 &
  56.92 &
  54.26 \\
 &
  DST\cite{papernot2016distillation} &
  80.96 &
  1.95 &
  15.76 &
  8.72 &
  11.31 &
  11.44 &
  10.32 &
  28.02 &
  27.54 &
  46.15 &
  48.28 &
  6.16 &
  64.94 &
  27.81 \\
 &
  HGD\cite{liao2018defense} &
  61.65 &
  45.48 &
  51.52 &
  52.73 &
  40.22 &
  36.96 &
  39.23 &
  51.52 &
  59.08 &
  61.28 &
  63.79 &
  51.24 &
  55.89 &
  51.58 \\
 &
  RT$^{*}$\cite{xie2017mitigating} &
  46.77 &
  21.44 &
  26.05 &
  16.28 &
  18.26 &
  22.24 &
  16.44 &
  13.95 &
  36.95 &
  44.10 &
  43.55 &
  24.42 &
  40.70 &
  28.55 \\
 &
  ST\cite{zheng2016improving} &
  70.69 &
  12.24 &
  12.24 &
  10.56 &
  17.05 &
  15.39 &
  13.81 &
  9.63 &
  59.29 &
  \textbf{67.83} &
  62.92 &
  15.11 &
  60.34 &
  32.85 \\
 &
  APE\cite{shen2017ape} &
  60.88 &
  65.01 &
  61.62 &
  65.80 &
  42.44 &
  43.60 &
  45.09 &
  14.63 &
  14.49 &
  54.6 &
  54.10 &
  30.91 &
  56.16 &
  46.87 \\
 &
  ARN\cite{zhou2021towards} &
  61.82 &
  61.34 &
  65.57 &
  65.04 &
  42.01 &
  40.65 &
  44.08 &
  60.45 &
  55.41 &
  60.17 &
  \textbf{70.12} &
  52.36 &
  68.57 &
  57.50 \\
\multirow{-9}{*}{\begin{tabular}[c]{@{}c@{}}CIFAR10\end{tabular}} &
  \textbf{MID} &
  64.48 &
  \textbf{69.88} &
  \textbf{66.47} &
  \textbf{68.08} &
  \textbf{45.12} &
  \textbf{41.63} &
  \textbf{45.09} &
  \textbf{62.24} &
  \textbf{59.39} &
  64.42 &
  64.48 &
  \textbf{53.29} &
  \textbf{65.70} &
  \textbf{59.25} \\ \hline
 &
  None &
  \textbf{94.90} &
  1.01 &
  7.56 &
  1.00 &
  6.75 &
  6.54 &
  8.18 &
  1.00 &
  7.92 &
  0.01 &
  10.00 &
  0.00 &
  9.99 &
  11.91 \\
 &
  $\rm AT'$\cite{madry2017towards} &
  71.98 &
  48.71 &
  57.58 &
  48.08 &
  38.98 &
  40.52 &
  44.24 &
  66.16 &
  53.79 &
  \textbf{60.44} &
  \textbf{60.22} &
  53.52 &
  38.37 &
  52.50 \\
\multirow{-3}{*}{\begin{tabular}[c]{@{}c@{}} SVHN\end{tabular}} &
  \textbf{MID} &
  66.72 &
  \textbf{59.18} &
  \textbf{65.31} &
  \textbf{69.80} &
  \textbf{67.24} &
  \textbf{64.84} &
  \textbf{67.97} &
  \textbf{70.04} &
  \textbf{75.21} &
  57.19 &
  56.09 &
  \textbf{56.28} &
  \textbf{43.98} &
  \textbf{63.06} \\ \hline
 &
  None &
  \textbf{60.36} &
  1.87 &
  2.87 &
  0.94 &
  1.09 &
  1.25 &
  1.15 &
  6.55 &
  22.92 &
  11.62 &
  23.75 &
  0.00 &
  1.00 &
  10.41 \\
 &
  $\rm AT'$\cite{madry2017towards} &
  37.32 &
  37.49 &
  31.56 &
  19.35 &
  11.17 &
  8.36 &
  6.20 &
  23.31 &
  32.04 &
  27.27 &
  27.31 &
  18.50 &
  30.51 &
  23.87 \\
 &
  HGD\cite{liao2018defense} &
  35.61 &
  36.15 &
  32.09 &
  28.74 &
  13.08 &
  12.92 &
  11.28 &
  29.63 &
  31.97 &
  25.91 &
  30.87 &
  19.66 &
  30.86 &
  26.05 \\
 &
  APE\cite{shen2017ape} &
  30.79 &
  29.27 &
  28.41 &
  21.81 &
  20.19 &
  15.01 &
  10.27 &
  30.18 &
  32.04 &
  25.09 &
  28.86 &
  18.63 &
  29.73 &
  24.63 \\
 &
  ARN\cite{zhou2021towards} &
  35.08 &
  37.98 &
  36.81 &
  29.69 &
  22.17 &
  16.39 &
  16.08 &
  29.79 &
  32.54 &
  31.31 &
  \textbf{38.17} &
  24.59 &
  30.71 &
  29.33 \\
\multirow{-6}{*}{\begin{tabular}[c]{@{}c@{}}CIFAR100\end{tabular}} &
  \textbf{MID} &
  31.30 &
  \textbf{38.15} &
  \textbf{38.66} &
  \textbf{30.96} &
  \textbf{24.73} &
  \textbf{18.60} &
  \textbf{16.12} &
  \textbf{33.14} &
  \textbf{33.97} &
  \textbf{31.48} &
  31.30 &
  \textbf{28.43} &
  \textbf{31.29} &
  \textbf{29.85} \\ \hline
    &
  None &
  53.70 &
  1.04 &
  0.90 &
  1.70 &
  0.94 &
  2.92 &
  2.72 &
  1.52 &
  0.96 &
  0.04 &
  0.05 &
  0.00 &
  1.25 &
  5.21 \\
 &
  $\rm AT'$\cite{madry2017towards} &
  40.84 &
  26.42 &
  27.52 &
  23.22 &
  22.44 &
  16.06 &
  23.88 &
  25.64 &
  21.02 &
  26.78 &
  25.71 &
  28.08 &
  12.19 &
  24.59 \\
\multirow{-3}{*}{\begin{tabular}[c]{@{}c@{}}Tiny-ImageNet\end{tabular}} &
  \textbf{MID} &
  38.56 &
  \textbf{28.00} &
  \textbf{28.78} &
  \textbf{23.66} &
  \textbf{24.90} &
  \textbf{22.91} &
  \textbf{25.72} &
  \textbf{26.64} &
  \textbf{23.30} &
  \textbf{27.61} &
  \textbf{26.78} &
  \textbf{29.22} &
  \textbf{16.64} &
  \textbf{26.36} \\ \hline
    &
  None &
  \textbf{60.38} &
  0.97 &
  1.03 &
  0.92 &
  3.74 &
  0.61 &
  0.53 &
  1.09 &
  3.75 &
  0.54 &
  1.64 &
  0.00 &
  1.78 &
  5.92 \\
 &
  $\rm AT'$\cite{madry2017towards} &
  40.16 &
  32.07 &
  30.79 &
  31.78 &
  32.87 &
  25.87 &
  26.02 &
  27.94 &
  30.87 &
  41.50 &
  \textbf{43.09} &
  30.32 &
  13.24  &
  31.27\\
\multirow{-3}{*}{\begin{tabular}[c]{@{}c@{}}ImageNet100\end{tabular}} &
  \textbf{MID} &
  41.90 &
  \textbf{36.74} &
  \textbf{34.67} &
  \textbf{33.92} &
  \textbf{32.97} &
  \textbf{33.68} &
  \textbf{34.49} &
  \textbf{34.44} &
  \textbf{33.71} &
  \textbf{42.90} &
  41.01 &
  \textbf{35.73} &
  \textbf{15.68}  &
  \textbf{34.75}\\ \hline
  &
  None &
  \textbf{81.91} &
  0.31 &
  0.16 &
  3.17 &
  0.84 &
  29.68 &
  33.06 &
  2.82 &
  1.04 &
  0.00 &
  6.91 &
  0.00 &
  - &
  13.32 \\
 &
  $\rm AT'$\cite{madry2017towards} &
  81.55 &
  41.52 &
  42.77 &
  53.68 &
  53.00 &
  47.10 &
  52.17 &
  54.82 &
  56.59 &
  54.16 &
  71.28 &
  51.09 &
  - &
  54.97 \\
\multirow{-3}{*}{\begin{tabular}[c]{@{}c@{}}ImageNet-1K\end{tabular}} &
  \textbf{MID}

  &79.56
  &\textbf{46.17}	&\textbf{48.61}	&\textbf{59.21}	&\textbf{67.13}	&\textbf{67.91}	&\textbf{69.88}	&\textbf{68.62}	&\textbf{71.03}	&\textbf{59.80}	&\textbf{72.24}	&\textbf{53.60}	&- &\textbf{63.64}\\ \hline
\end{tabular}}
\end{table*}

\textbf{Implementation Details. }
All experiments except ImageNet are run on 4 NVIDIA GTX2080ti with 11GB video memory, and on 8 NVIDIA RTX 4090 with 24GB memory for ImageNet. The batch size is 128 (except ImageNet) and 32 (ImageNet), and Pytorch version is 0.4.0. We use the implementation codes of 7 kinds of attacks, such as PGD \cite{madry2017towards}, MIM \cite{dong2018boosting}, FGSM \cite{goodfellow2014explaining}, BIM \cite{kurakin2016adversarial}, CW \cite{carlini2017towards}, JSMA \cite{papernot2016limitations} in the advertorch toolbox \cite{ding2019advertorch}, and Auto-Attack \cite{croce2020reliable} released by the developers. We set the maximum default disturbance $\epsilon$ = 0.3 and the initial learning rate is $lr$ = 1e-3.

In MID, the \emph{Attacker Pool} is manually fixed for training. The basic selection criterion is the number of known attacks should be small due to limited resource in real scenario and the known attacks can be representative. We adopt PGD and MIM as the known attack in the \emph{Attacker Pool} for training, and benign samples are naturally included in the $Attacker Pool$. Both PGD and MIM have target and non-target attack versions, which are expressed as PGD$_T$ for target PGD and PGD$_N$ for non-target PGD, and the same manipulation for other attacks. Therefore, 4 kinds of attacks are included in the \emph{Attacker Pool}. Notably, the simulated known attack (meta-train) and the simulated unknown attacks (meta-test) in training stage are sampled from the attacker pool. Attacks (generated by the released teacher model) are considered as static domains and are fixed to evaluate the attack invariant information during testing. FGSM, BIM, CW, AA and JSMA, as the real unknown attacks, are only used in test phase.

For the white box attack, we directly attack the target model. For black box attack, we adopt the commonly used attack strategy: we train a new alternative model for each dataset, which has the same network structure and loss function as the real target model trained by the same datasets. Such transfer-based strategy for black box attack has been proved to be effective. Notably, for targeted attacks (e.g., $\rm PGD_T$) on MNIST and CIFAR10 consisting of 10 categories, the target label is randomly selected, so the attacked accuracy for the baseline model shall be close to 10\% where all samples are attacked to the same target label.


\begin{table*}[]
\caption{Defense for commonly evaluated black box attacks. Higher accuracy means better robustness. The best robustness is marked in bold.}
\vspace{-0.2cm}
\label{tab:2}
\scriptsize
\centering
\setlength{\tabcolsep}{1.4mm}{
\begin{tabular}{c|c|c|cccc|cccccccc|c}
\hline
   &
   &
   &
  \multicolumn{4}{c|}{Known Attack} &
  \multicolumn{8}{c|}{Unknown Attack} \\ \cline{4-15}
\multirow{-2}{*}{} &
  \multirow{-2}{*}{Defense} &
  \multirow{-2}{*}{Benign} &
  $\rm PGD_N$ &
  $\rm PGD_T$ &
  $\rm MIM_N$ &
  $\rm MIM_T$ &
  $\rm FGSM_N$ &
  $\rm FGSM_T$ &
  $\rm BIM_N$ &
  $\rm BIM_T$ &
  $\rm CW_N$ &
  $\rm CW_T$ &
  $\rm AA_N$ &
  $\rm JSMA_T$ &
  \multirow{-2}{*}{Avg.}
  \\ \hline
 &
  None &
  \textbf{99.16} &
  12.09 &
  54.06 &
  40.48 &
  22.37 &
  57.52 &
  44.73 &
  1.18 &
  88.49 &
  96.95 &
  96.16 &
  5.09 &
  28.19 &
  49.72 \\
 &
  $\rm AT'$\cite{madry2017towards} &
  98.70 &
  97.23 &
  77.58 &
  96.47 &
  91.49 &
  90.99 &
  88.02 &
  93.94 &
  83.52 &
  97.29 &
  98.20 &
  96.63 &
  51.16 &
  89.32 \\
 &
  DST\cite{papernot2016distillation} &
  98.93 &
  58.97 &
  67.07 &
  43.72 &
  39.60 &
  58.59 &
  45.50 &
  40.38 &
  88.20 &
  97.24 &
  96.21 &
  10.54 &
  65.15  &
  62.31\\
 &
  DOA\cite{wu2019defending} &
  91.73 &
  35.78 &
  69.89 &
  68.65 &
  73.84 &
  83.79 &
  83.95 &
  81.08 &
  83.17 &
  89.56 &
  90.52 &
  10.59 &
  76.91 &
  72.26 \\
 &
  HGD\cite{liao2018defense} &
  90.4 &
  93.05 &
  92.51 &
  84.46 &
  85.98 &
  90.17 &
  90.12 &
  92.47 &
  91.92 &
  95.54 &
  96.97 &
  94.97 &
  42.98 &
  87.81 \\
 &
  RT$^{*}$\cite{xie2017mitigating} &
  84.57 &
  27.93 &
  74.20 &
  41.67 &
  40.06 &
  57.60 &
  43.45 &
  35.50 &
  89.65 &
  91.60 &
  89.40 &
  20.26 &
  52.60 &
  57.57 \\
 &
  ST\cite{zheng2016improving} &
  99.21 &
  67.07 &
  81.51 &
  55.17 &
  46.50 &
  62.24 &
  47.59 &
  39.37 &
  93.99 &
  97.71 &
  98.01 &
  15.33 &
  72.49 &
  67.39 \\
 &
  APE\cite{shen2017ape} &
  99.05 &
  93.33 &
  96.32 &
  89.33 &
  93.43 &
  86.25 &
  80.15 &
  87.17 &
  92.89 &
  94.13 &
  94.68 &
  91.25 &
  48.53 &
  88.19 \\
 &
  ARN\cite{zhou2021towards} &
  91.79 &
  97.16 &
  97.68 &
  92.41 &
  93.98 &
  91.56 &
  90.01 &
  94.58 &
  96.00 &
  97.56 &
  98.08 &
  96.58 &
  \textbf{69.52} &
  92.83 \\
\multirow{-10}{*}{\begin{tabular}[c]{@{}c@{}}MNIST\end{tabular}} &
  \textbf{MID} &
  98.83 &
  \textbf{97.32} &
  \textbf{98.06} &
  \textbf{96.66} &
  \textbf{94.03} &
  \textbf{92.40} &
  \textbf{89.95} &
  \textbf{97.39} &
  \textbf{96.38} &
  \textbf{98.28} &
  \textbf{98.81} &
  \textbf{97.54} &
  53.59 &
  \textbf{93.01} \\ \hline
 &
  None &
  \textbf{91.81} &
  6.27 &
  58.80 &
  31.00 &
  15.82 &
  37.61 &
  19.37 &
  30.99 &
  40.71 &
  68.68 &
  67.84 &
  3.61 &
  28.24 &
  38.51 \\
 &
  $\rm AT'$\cite{madry2017towards} &
  91.62 &
  84.68 &
  65.94 &
  84.04 &
  57.62 &
  57.87 &
  55.31 &
  86.82 &
  68.75 &
  86.63 &
  89.08 &
  88.53 &
  49.45 &
  74.33 \\
 &
  HGD\cite{liao2018defense} &
  81.07 &
  82.17 &
  81.67 &
  81.93 &
  75.96 &
  71.93 &
  72.52 &
  81.54 &
  82.88 &
  76.21 &
  79.57 &
  85.93 &
  40.78 &
  76.47 \\
 &
  APE\cite{shen2017ape} &
  89.07 &
  79.34 &
  82.74 &
  81.83 &
  67.21 &
  69.04 &
  70.29 &
  80.54 &
  81.39 &
  83.80 &
  84.12 &
  83.93 &
  41.02 &
  76.48 \\
 &
  ARN\cite{zhou2021towards} &
  90.36 &
  81.36 &
  83.29 &
  82.22 &
  82.19 &
  71.37 &
  74.85 &
  82.72 &
  84.43 &
  84.29 &
  86.17 &
  86.73 &
  \textbf{55.71} &
  80.43 \\
\multirow{-6}{*}{\begin{tabular}[c]{@{}c@{}}Fashion \\ MNIST\end{tabular}} &
  \textbf{MID} &
  89.75 &
  \textbf{86.72} &
  \textbf{86.15} &
  \textbf{87.37} &
  \textbf{79.57} &
  \textbf{74.32} &
  \textbf{79.02} &
  \textbf{87.78} &
  \textbf{87.17} &
  \textbf{89.37} &
  \textbf{89.70} &
  \textbf{89.46} &
  41.57 &
  \textbf{82.14} \\ \hline
 &
  None &
  \textbf{83.99} &
  1.55 &
  15.01 &
  8.62 &
  11.84 &
  17.87 &
  13.85 &
  6.50 &
  31.90 &
  78.68 &
  76.90 &
  6.09 &
  17.09 &
  28.45 \\
 &
  $\rm AT'$\cite{madry2017towards} &
  63.24 &
  13.88 &
  28.25 &
  13.91 &
  13.37 &
  13.78 &
  13.45 &
  17.33 &
  40.98 &
  58.28 &
  56.68 &
  21.22 &
  59.25 &
  31.81 \\
 &
  DST\cite{papernot2016distillation} &
  80.96 &
  1.49 &
  16.40 &
  7.71 &
  10.90 &
  10.91 &
  10.32 &
  5.93 &
  55.53 &
  \textbf{67.18} &
  \textbf{67.23} &
  6.04 &
  65.79 &
  31.26 \\
 &
  DOA\cite{wu2019defending} &
  80.18 &
  8.69 &
  24.65 &
  43.69 &
  35.50 &
  \textbf{36.67} &
  32.54 &
  42.88 &
  53.16 &
  50.49 &
  58.68 &
  3.47 &
  65.79 &
  41.26 \\
 &
  HGD\cite{liao2018defense} &
  61.65 &
  42.13 &
  49.98 &
  51.09 &
  50.01 &
  31.84 &
  32.96 &
  50.51 &
  51.08 &
  62.93 &
  64.08 &
  46.89 &
  56.01 &
  50.08 \\
 &
  RT$^{*}$\cite{xie2017mitigating} &
  46.77 &
  12.76 &
  25.93 &
  14.02 &
  17.83 &
  16.34 &
  16.01 &
  11.80 &
  37.75 &
  46.30 &
  45.60 &
  13.85 &
  43.10 &
  26.77 \\
 &
  ST\cite{zheng2016improving} &
  70.69 &
  1.76 &
  50.49 &
  9.41 &
  17.49 &
  14.40 &
  14.65 &
  8.16 &
  52.13 &
  67.70 &
  66.64 &
  5.07 &
  60.34 &
  33.76 \\
 &
  APE\cite{shen2017ape} &
  60.88 &
  56.34 &
  55.98 &
  43.93 &
  33.37 &
  33.78 &
  33.45 &
  14.66 &
  14.91 &
  58.28 &
  56.68 &
  50.25 &
  44.50 &
  42.84 \\
 &
  ARN\cite{zhou2021towards} &
  61.82 &
  56.67 &
  62.38 &
  46.56 &
  30.71 &
  30.96 &
  31.78 &
  59.68 &
  62.76 &
  65.64 &
  61.66 &
  50.16 &
  64.89 &
  52.74 \\
\multirow{-10}{*}{\begin{tabular}[c]{@{}c@{}}CIFAR10\end{tabular}} &
  \textbf{MID} &
  64.48 &
  \textbf{64.84} &
  \textbf{65.98} &
  \textbf{52.69} &
  \textbf{38.97} &
  \textbf{34.41} &
  \textbf{36.25} &
  \textbf{62.94} &
  \textbf{66.06} &
  64.34 &
  64.48 &
  \textbf{52.02} &
  \textbf{60.88} &
  \textbf{56.02} \\ \hline
 &
  None &
  \textbf{94.90} &
  0.21 &
  7.07 &
  2.63 &
  11.95 &
  12.81 &
  10.35 &
  2.60 &
  12.75 &
  64.75 &
  62.57 &
  0.07 &
  19.64 &
  23.25 \\
 &
  $\rm AT'$\cite{madry2017towards} &
  71.98 &
  59.47 &
  53.79 &
  10.38 &
  27.45 &
  43.55 &
  46.82 &
  43.20 &
  42.26 &
  72.13 &
  70.55 &
  42.15 &
  30.08  &
  47.21\\
  \multirow{-3}{*}{\begin{tabular}[c]{@{}c@{}}SVHN\end{tabular}} &
  \textbf{MID} &
  66.72 &
  \textbf{65.84} &
  \textbf{68.04} &
  \textbf{70.52} &
  \textbf{76.93} &
  \textbf{58.95} &
  \textbf{64.58} &
  \textbf{60.58} &
  \textbf{65.68} &
  \textbf{62.50} &
  \textbf{62.54} &
  \textbf{52.20} &
  47.01 &
  \textbf{63.23} \\ \hline
 &
  None &
  \textbf{60.36} &
  1.53 &
  5.81 &
  10.33 &
  2.03 &
  1.26 &
  1.10 &
  5.77 &
  26.45 &
  17.65 &
  20.30 &
  11.19 &
  11.96 &
  13.51 \\
 &
  $\rm AT'$\cite{madry2017towards} &
  37.32 &
  24.03 &
  30.36 &
  16.42 &
  9.46 &
  8.20 &
  5.72 &
  15.98 &
  20.24 &
  37.30 &
  23.32 &
  16.68 &
  28.12 &
  21.01 \\
 &
  HGD\cite{liao2018defense} &
  35.61 &
  18.54 &
  16.93 &
  12.85 &
  11.04 &
  10.88 &
  10.25 &
  19.38 &
  25.65 &
  18.04 &
  21.71 &
  15.18 &
  28.57 &
  18.81 \\
 &
  APE\cite{shen2017ape} &
  30.79 &
  19.21 &
  17.49 &
  15.81 &
  15.19 &
  11.43 &
  11..27 &
  21.16 &
  23.05 &
  22.01 &
  23.38 &
  17.63 &
  24.92 &
  19.48 \\
 &
  ARN\cite{zhou2021towards} &
  35.08 &
  24.98 &
  28.81 &
  18.72 &
  16.83 &
  13.89 &
  10.62 &
  22.08 &
  28.07 &
  24.37 &
  \textbf{26.71} &
  29.98 &
  18.49 &
  22.97 \\
\multirow{-6}{*}{\begin{tabular}[c]{@{}c@{}}CIFAR100\end{tabular}} &
  \textbf{MID} &
  31.30 &
  \textbf{26.55} &
  \textbf{30.80} &
  \textbf{24.41} &
  \textbf{17.71} &
  \textbf{14.92} &
  \textbf{12.15} &
  \textbf{22.61} &
  \textbf{32.87} &
  \textbf{25.67} &
  25.64 &
  \textbf{30.64} &
  \textbf{20.59} &
  \textbf{24.29} \\ \hline
  &
  None &
  \textbf{53.70} &
  3.00 &
  4.84 &
  3.10 &
  3.86 &
  3.66 &
  3.48 &
  2.76 &
  4.58 &
  20.31 &
  18.35 &
  16.54 &
  4.06 &
  10.94 \\
 &
  $\rm AT'$\cite{madry2017towards} &
  40.84 &
  23.96 &
  26.62 &
  19.68 &
  21.24 &
  13.96 &
  13.98 &
  21.28 &
  \textbf{26.16} &
  26.27 &
  24.20 &
  25.72 &
  17.50  &
  23.18\\
\multirow{-3}{*}{\begin{tabular}[c]{@{}c@{}}Tiny-ImageNet\end{tabular}} &
  \textbf{MID} &
  38.56 &
  \textbf{24.44} &
  \textbf{28.24} &
  \textbf{22.28} &
  \textbf{23.84} &
  \textbf{19.06} &
  \textbf{18.96} &
  \textbf{23.50} &
  25.76 &
  \textbf{27.04} &
  26.86 &
  \textbf{25.54} &
  \textbf{18.12} &
  \textbf{24.78} \\ \hline
    &
  None &
  \textbf{60.38} &
  8.21 &
  7.97 &
  6.92 &
  10.21 &
  9.31 &
  12.74 &
  7.93 &
  10.90 &
  10.04 &
  12.05 &
  8.97 &
  10.21 &
  13.52 \\
 &
  $\rm AT'$\cite{madry2017towards} &
  40.16 &
  30.87 &
  28.46 &
  28.07 &
  28.89 &
  25.62 &
  26.98 &
  26.82 &
  29.39 &
  \textbf{42.56} &
  41.82 &
  30.01 &
  13.73 &
  30.25 \\
\multirow{-3}{*}{\begin{tabular}[c]{@{}c@{}}ImageNet100\end{tabular}} &
  \textbf{MID} &
  41.90 &
  \textbf{35.22} &
  \textbf{32.87} &
  \textbf{30.81} &
  \textbf{31.78} &
  \textbf{33.34} &
  \textbf{32.07} &
  \textbf{34.84} &
  \textbf{32.10} &
  41.98 &
  \textbf{42.20} &
  \textbf{32.78} &
  \textbf{15.80} &
  \textbf{33.66} \\ \hline
  &
  None
  &\textbf{81.91}	&21.36	&20.64	&20.14	&29.05	&29.05	&35.14	&25.44	&28.81	&19.07	&23.21	&20.16	&- & 29.49

  \\
 &
  $\rm AT'$\cite{madry2017towards}
    &81.55	&39.20	&40.21	&48.56	&55.45	&48.06	&47.94	&68.30	&68.91	&55.04	&70.86	&50.84	&- &56.24
  \\
\multirow{-3}{*}{\begin{tabular}[c]{@{}c@{}}ImageNet-1K\end{tabular}} &
  \textbf{MID}
    &79.56	&\textbf{48.81}	&\textbf{49.85}	&\textbf{58.58}	&\textbf{66.58}	&\textbf{67.50}	&\textbf{68.62}	&\textbf{68.65}	&\textbf{71.29}	&\textbf{59.12}	&\textbf{71.96}	&\textbf{52.91}	&- &\textbf{63.61}
  \\ \hline
\end{tabular}}
\vspace{-0.2cm}
\end{table*}

\subsection{Main Results of MID}
\textbf{Generalizable Robustness}. The performance w.r.t. known and unknown attacks is evaluated on white-box and black-box attacks, respectively, which are presented in Tab.\ref{tab:1} and Tab.\ref{tab:2}. 
Notably, for black-box attacks, we train a substitute model which has the same structure as the target model.

\begin{table}[]
\caption{Black-box attacks from different source models on MNIST.}
\centering
\setlength{\tabcolsep}{2mm}{
\label{tab:black_mnist}
\begin{tabular}{c|cccccc}
\hline
\multirow{2}{*}{MNIST} & \multicolumn{2}{c}{LeNet5} & \multicolumn{2}{c}{ResNet18} & \multicolumn{2}{c}{ResNet50} \\ \cline{2-7}
      &  AT & \textbf{MID}            & AT     & \textbf{MID}            & AT     & \textbf{MID}            \\ \hline
PGD$\rm _N$  & 97.23      & \textbf{97.32} & 94.14          & \textbf{93.82} & 94.28           & \textbf{94.73} \\
PGD$\rm _T$  & 77.58      & \textbf{98.06} & 84.81          & \textbf{94.81} & 84.07          & \textbf{94.13} \\
MIM$\rm _N$  & 96.47      & \textbf{96.66} & 92.14          & \textbf{95.87} & 90.44          & \textbf{92.20} \\
MIM$\rm _T$  & 91.49      & \textbf{94.03} & 92.68          & \textbf{94.60} & 90.81          & \textbf{92.16} \\
FGSM$\rm _N$ & 90.99      & \textbf{92.40} & 88.13          & \textbf{90.59} & 90.81          & \textbf{91.36} \\
FGSM$\rm _T$ & 88.02      & \textbf{89.95} & 89.20          & \textbf{90.19} & \textbf{90.45} & 90.23          \\
BIM$\rm _N$  & 93.94      & \textbf{97.39} & 88.56          & \textbf{91.61} & 88.79          & \textbf{89.50} \\
BIM$\rm _T$  & 83.52      & \textbf{96.38} & 90.49          & \textbf{93.12} & 88.16          & \textbf{89.22} \\
CW$\rm _N$   & 79.29      & \textbf{98.28} & \textbf{98.50} & 98.46          & \textbf{98.47} & 98.40          \\
CW$\rm _T$   & 98.20      & \textbf{98.81} & 93.46          & \textbf{93.49} & 96.17          & \textbf{98.26} \\
AA$\rm _N$   & 96.63      & \textbf{97.54} & 92.17          & \textbf{95.10} & 92.29          & \textbf{93.71} \\ \hline
\end{tabular}}
\vspace{-0.2cm}
\end{table}

\begin{table*}[]
\caption{Black box attacks from different source models on CIFAR10.}
\centering
\scriptsize
\setlength{\tabcolsep}{3mm}{
\label{tab:black_cifar10}
\begin{tabular}{c|cccccccccccc}
\hline
\multirow{2}{*}{CIFAR10} & \multicolumn{3}{c}{ResNet18} & \multicolumn{3}{c}{ResNet50} & \multicolumn{3}{c}{DenseNet169} & \multicolumn{3}{c}{ConvNext-B} \\ \cline{2-13}
 &
  AT &
  RSLAD &
  \textbf{MID} &
  AT &
  RSLAD &
  \textbf{MID} &
  AT &
  RSLAD &
  \textbf{MID} &
  AT &
  RSLAD &
  \textbf{MID} \\ \hline
PGD$\rm _N$ &
  13.88 &
  31.95 &
  \textbf{64.84} &
  25.61 &
  35.39 &
  \textbf{54.74} &
  8.43 &
  30.21 &
  \textbf{41.99} &
  23.48 &
  41.17 &
  \textbf{52.93} \\
PGD$\rm _T$ &
  28.25 &
  52.58 &
  \textbf{65.98} &
  27.94 &
  55.18 &
  \textbf{66.11} &
  25.80 &
  52.89 &
  \textbf{64.53} &
  29.10 &
  51.64 &
  \textbf{64.00} \\
MIM$\rm _N$ &
  13.91 &
  29.22 &
  \textbf{52.69} &
  29.95 &
  30.47 &
  \textbf{36.47} &
  11.27 &
  27.07 &
  \textbf{28.12} &
  18.01 &
  30.33 &
  \textbf{24.41} \\
MIM$\rm _T$ &
  13.37 &
  15.70 &
  \textbf{38.97} &
  14.51 &
  21.27 &
  \textbf{41.34} &
  14.84 &
  24.77 &
  \textbf{36.47} &
  25.64 &
  32.03 &
  \textbf{32.15} \\
FGSM$\rm _N$ &
  13.78 &
  13.98 &
  \textbf{34.41} &
  14.33 &
  16.60 &
  \textbf{32.47} &
  13.15 &
  17.83 &
  \textbf{31.99} &
  35.59 &
  43.20 &
  \textbf{47.40} \\
FGSM$\rm _T$ &
  13.45 &
  14.30 &
  \textbf{36.25} &
  13.33 &
  14.92 &
  \textbf{32.49} &
  11.49 &
  16.00 &
  \textbf{24.05} &
  34.71 &
  41.88 &
  \textbf{48.27} \\
BIM$\rm _N$ &
  17.33 &
  30.55 &
  \textbf{62.94} &
  29.62 &
  31.23 &
  \textbf{36.47} &
  10.41 &
  25.30 &
  \textbf{25.66} &
  39.88 &
  32.66 &
  \textbf{42.24} \\
BIM$\rm _T$ &
  40.98 &
  36.88 &
  \textbf{66.06} &
  24.46 &
  30.20 &
  \textbf{63.20} &
  19.82 &
  38.81 &
  \textbf{55.28} &
  34.92 &
  39.69 &
  \textbf{44.98} \\
CW$\rm _N$ &
  58.28 &
  \textbf{68.28} &
  64.34 &
  57.85 &
  \textbf{69.12} &
  66.55 &
  58.48 &
  \textbf{68.12} &
  61.34 &
  60.55 &
  63.05 &
  \textbf{66.53} \\
CW$\rm _T$ &
  56.68 &
  \textbf{68.12} &
  64.48 &
  51.67 &
  66.54 &
  \textbf{66.86} &
  55.61 &
  66.41 &
  \textbf{66.86} &
  56.16 &
  60.61 &
  \textbf{66.51} \\
AA$\rm _N$ &
  21.22 &
  21.48 &
  \textbf{52.02} &
  6.27 &
  21.70 &
  \textbf{43.58} &
  7.11 &
  18.48 &
  \textbf{23.56} &
  30.31 &
  36.72 &
  \textbf{47.02} \\ \hline
\end{tabular}}
\end{table*}

\begin{table*}[]
\caption{Black box attacks from different source models on CIFAR100.}
\centering
\scriptsize
\setlength{\tabcolsep}{3mm}{
\label{tab:black_cifar100}
\begin{tabular}{c|cccccccccccc}
\hline
\multirow{2}{*}{CIFAR100} &
  \multicolumn{3}{c}{ResNet18} &
  \multicolumn{3}{c}{ResNet50} &
  \multicolumn{3}{c}{DenseNet169} &
  \multicolumn{3}{c}{ConvNext-B} \\ \cline{2-13}
      & AT    & RSLAD & \textbf{MID}            & AT    & RSLAD & \textbf{MID}            & AT    & RSLAD & \textbf{MID}            & AT    & RSLAD & \textbf{MID}            \\ \hline
PGD$\rm _N$  & 24.03 & 26.50 & \textbf{26.55} & 17.93 & 16.15 & \textbf{25.12} & 21.50 & 12.19 & \textbf{23.61} & 37.31 & 12.40 & \textbf{38.41} \\
PGD$\rm _T$  & 30.36 & 29.30 & \textbf{30.80} & 30.94 & 17.92 & \textbf{31.61} & 32.74 & 23.23 & \textbf{34.67} & 27.65 & 17.29 & \textbf{29.87} \\
MIM$\rm _N$  & 16.42 & 11.06 & \textbf{24.41} & 12.13 & 13.10 & \textbf{24.09} & 17.23 & 13.44 & \textbf{20.86} & 16.47 & 15.10 & \textbf{22.46} \\
MIM$\rm _T$  & 9.46  & 6.22  & \textbf{17.71} & 12.52 & 15.00 & \textbf{15.89} & 16.45 & 11.35 & \textbf{18.81} & 14.93 & 12.92 & \textbf{20.62} \\
FGSM$\rm _N$ & 8.20  & 13.78 & \textbf{14.92} & 12.09 & 13.52 & \textbf{14.06} & 8.97  & 12.29 & \textbf{13.69} & 14.94 & 13.12 & \textbf{15.61} \\
FGSM$\rm _T$ & 5.72  & 13.12 & \textbf{12.15} & 11.25 & 12.84 & \textbf{12.00} & 8.38  & 10.19 & \textbf{11.52} & 14.64 & 12.10 & \textbf{16.00} \\
BIM$\rm _N$  & 15.98 & 19.71 & \textbf{22.61} & 13.48 & 12.19 & \textbf{21.61} & 17.52 & 18.44 & \textbf{21.34} & 11.47 & 18.96 & \textbf{20.72} \\
BIM$\rm _T$  & 20.24 & 16.15 & \textbf{32.87} & 19.41 & 14.01 & \textbf{30.49} & 21.60 & 26.04 & \textbf{31.57} & 13.90 & 12.29 & \textbf{32.02} \\
CW$\rm _N$   &
  27.30 &
  \textbf{26.17} &
  25.67 &
  26.01 &
  \textbf{23.82} &
  \textbf{26.97} &
  \textbf{27.32} &
  \textbf{27.29} &
  26.88 &
  25.08 &
  25.60 &
  \textbf{25.68} \\
CW$\rm _T$ &
  23.32 &
  \textbf{25.12} &
  \textbf{25.64} &
  \textbf{26.79} &
  24.17 &
  \textbf{26.61} &
  25.42 &
  25.09 &
  \textbf{26.06} &
  25.14 &
  \textbf{27.29} &
  \textbf{25.67} \\
AA$\rm _N$   & 16.68 & 23.80 & \textbf{30.64} & 16.91 & 17.16 & \textbf{31.62} & 16.43 & 28.44 & \textbf{32.64} & 19.94 & 17.40 & \textbf{33.61} \\ \hline
\end{tabular}}
\end{table*}

Apparently, MID always shows the best robustness to both the known and unknown attacks and achieves the state-of-the-art performance. For the known attacks used in the training stage (i.e., PGD and MIM), MID can always maintain the best robustness. For the unknown attacks (i.e., FGSM, BIM, CW, AA and JSMA), we find that the robustness of MID on CW does not always reach the best performance. This may be because CW is an optimization-based attack method while several other attacks are gradient-based. Therefore, the invariant information between the those attacks may be limited. But MID still achieves the best robustness against most unknown attacks.

We further discuss black-box attacks from different source models in Tab. \ref{tab:black_mnist}, Tab. \ref{tab:black_cifar10} and Tab. \ref{tab:black_cifar100} for MNIST, CIFAR10 and CIFAR100, respectively. We compare MID with adversarial training and defensive distillation models. We observe that for different source models, MID always performs better than others. Besides, we discuss more unknown attacks, and evaluate the performance of MID against several heuristic or advanced attacks (i.e., SPA \cite{su2019one}, SSAH \cite{luo2022frequency}, stAdv \cite{xiao2018spatially}, DDN \cite{rony2019decoupling}, FAB \cite{croce2020minimally} and Adv-Drop \cite{duan2021advdrop}) on CIFAR10, CIFAR100 and Tiny-ImageNet in Tab. \ref{tab:advance}, which show the superiority of MID.

From the results in Tab.\ref{tab:1} and Tab.\ref{tab:2}, we see that although the accuracy of MID for benign samples is sometimes slightly lower than other defense models, the average robustness (\textbf{Avg.}) of MID is always the best. Besides, the worst-case adversarial robustness (i.e., the lowest performance) of MID is always stronger than other defense models, which means MID not only has a higher upper bound and expectation, but also a better lower bound compared with others. Therefore, we believe MID remains the best robustness for adversarial samples and average robustness, despite the slight degradation for benign samples.

\begin{table}[]
\caption{Test on several heuristic or advanced unknown attacks. 
}
\centering
 \scriptsize
\setlength{\tabcolsep}{1mm}{
\label{tab:advance}
\begin{tabular}{c|c|cccccc}
\hline
Datasets                      & Defense    & SPA   & SSAH  & stAdv & DDN   & FAB   & Adv-Drop \\ \hline
\multirow{3}{*}{CIFAR10}      & Baseline   & 20.05 & 0.58  & 9.77  & 9.77  & 9.69  & 10.78    \\
                              & AT         & 63.75 & 62.19 & 47.50 & 52.50 & 52.27 & 12.58    \\
                              & \textbf{MID}        & \textbf{65.62} & \textbf{64.64} & \textbf{50.41} & \textbf{54.71} & \textbf{55.37} & \textbf{16.31}    \\ \hline
\multirow{3}{*}{CIFAR100}     & Baseline   & 14.38 & 7.54  & 12.40 & 12.50 & 12.50 & 1.77     \\
                              & AT         & 27.50 & 21.25 & 26.06 & 27.92 & 27.65 & 20.94    \\
                              & \textbf{MID}        & \textbf{28.15} & \textbf{24.04} & \textbf{28.12} & \textbf{28.50} & \textbf{28.54} & \textbf{22.88}    \\ \hline
\multirow{3}{*}{\begin{tabular}[c]{@{}c@{}}Tiny\\ ImageNet\end{tabular}}
                              & Baseline   & 18.30 & 0.52  & 9.84  & 0.52  & 8.20  & 1.56     \\
                              & AT         & 39.26 & 17.49 & 21.55 & 21.86 & 20.07 & 24.50    \\
                              & \textbf{MID}        & \textbf{33.89} & \textbf{20.49} & \textbf{25.57} & \textbf{22.95} & \textbf{24.22} & \textbf{26.88}    \\ \hline
\end{tabular}}
\vspace{-0.1cm}
\end{table}

\begin{table*}[]
\caption{Ablation analysis. Meta and DST mean the meta-learning and distillation framework respectively. AC and CC mean the Adversarial- and Cyclic Consistency constraints of MID respectively. The Label Consistency is naturally contained in each module. The best robustness is marked in bold.}
\centering
\scriptsize
\setlength{\tabcolsep}{1.3mm}{
\label{tab:abla}
\begin{tabular}{c|cccc|c|cccc|ccccccc}
\hline
\multirow{2}{*}{Datasets} &
  \multicolumn{4}{c|}{Modules} &
  \multirow{2}{*}{Clean} &
  \multicolumn{4}{c|}{Known Attacks} &
  \multicolumn{7}{c}{Unknown attacks} \\ \cline{2-5} \cline{7-17}
 &
  Meta &
  DST &
  AC &
  CC &
   &
  $\rm PGD_N$ &
  $\rm PGD_T$ &
  $\rm MIM_N$ &
  $\rm MIM_T$ &
  $\rm FGSM_N$ &
  $\rm FGSM_T$ &
  $\rm BIM_N$ &
  $\rm BIM_T$ &
  $\rm CW_N$ &
  $\rm CW_T$ &
  $\rm AA_N$ \\ \hline
\multirow{6}{*}{MNIST} &
   &
   &
   &
   &
  \textbf{99.04} &
  99.00 &
  98.81 &
  98.11 &
  97.82 &
  91.88 &
  92.36 &
  93.14 &
  88.96 &
  95.91 &
  95.69 &
  96.45 \\
 &
  \checkmark &
   &
   &
   &
  98.14 &
  98.13 &
  98.02 &
  98.09 &
  98.80 &
  93.01 &
  95.55 &
  95.08 &
  90.64 &
  96.04 &
  97.98 &
  97.51 \\
 &
  \checkmark &
  \checkmark &
   &
   &
  98.64 &
  98.99 &
  98.24 &
  98.64 &
  98.02 &
  93.38 &
  95.76 &
  96.54 &
  89.09 &
  97.59 &
  98.59 &
  98.15 \\
 &
  \checkmark &
  \checkmark &
  \checkmark &
   &
  98.66 &
  98.99 &
  98.24 &
  99.00 &
  98.10 &
  93.41 &
  96.71 &
  97.81 &
  97.07 &
  98.21 &
  98.60 &
  98.20 \\
 &
  \checkmark &
  \checkmark &
   &
  \checkmark &
  98.57 &
  99.00 &
  98.24 &
  98.56 &
  98.08 &
  93.42 &
  96.68 &
  98.01 &
  97.99 &
  \textbf{98.47} &
  98.54 &
  98.20 \\
 &
  \checkmark &
  \checkmark &
  \checkmark &
  \checkmark &
  98.83 &
  \textbf{99.01} &
  \textbf{99.20} &
  \textbf{99.05} &
  \textbf{98.15} &
  \textbf{93.47} &
  \textbf{96.72} &
  \textbf{98.90} &
  \textbf{98.78} &
  98.30 &
  \textbf{98.70} &
  \textbf{98.24} \\ \hline
\multirow{6}{*}{CIFAR10} &
   &
   &
   &
   &
  65.77 &
  68.91 &
  65.16 &
  62.08 &
  43.50 &
  34.59 &
  41.80 &
  59.49 &
  50.11 &
  61.21 &
  41.81 &
  50.24 \\
 &
  \checkmark &
   &
   &
   &
  66.73 &
  68.10 &
  65.76 &
  63.37 &
  43.22 &
  37.17 &
  43.55 &
  60.20 &
  52.55 &
  63.82 &
  64.57 &
  52.43 \\
 &
  \checkmark &
  \checkmark &
   &
   &
  \textbf{66.81} &
  68.76 &
  65.27 &
  63.20 &
  41.29 &
  39.09 &
  43.30 &
  60.58 &
  53.18 &
  63.70 &
  64.31 &
  52.04 \\
 &
  \checkmark &
  \checkmark &
  \checkmark &
   &
  66.84 &
  69.37 &
  66.50 &
  63.53 &
  43.06 &
  39.33 &
  43.47 &
  61.20 &
  54.25 &
  64.69 &
  64.04 &
  52.29 \\
 &
  \checkmark &
  \checkmark &
   &
  \checkmark &
  66.72 &
  69.56 &
  66.70 &
  64.82 &
  43.02 &
  39.50 &
  43.41 &
  61.03 &
  59.39 &
  \textbf{65.60} &
  \textbf{66.72} &
  52.84 \\
 &
  \checkmark &
  \checkmark &
  \checkmark &
  \checkmark &
  64.48 &
  \textbf{69.88} &
  \textbf{66.47} &
  \textbf{65.08} &
  \textbf{45.12} &
  \textbf{41.63} &
  \textbf{45.09} &
  \textbf{62.24} &
  \textbf{59.39} &
  64.42 &
  64.48 &
  \textbf{53.29} \\ \hline
\multirow{6}{*}{CIFAR100} &
   &
   &
   &
   &
  32.34 &
  31.24 &
  37.77 &
  31.90 &
  19.13 &
  14.71 &
  12.85 &
  26.80 &
  30.84 &
  22.56 &
  22.23 &
  21.40 \\
 &
  \checkmark &
   &
   &
   &
  32.41 &
  35.29 &
  32.36 &
  35.27 &
  20.25 &
  15.61 &
  13.65 &
  30.78 &
  32.25 &
  26.29 &
  26.43 &
  23.00 \\
 &
  \checkmark &
  \checkmark &
   &
   &
  30.72 &
  37.55 &
  38.53 &
  29.45 &
  23.01 &
  16.18 &
  14.39 &
  31.85 &
  33.63 &
  28.85 &
  29.63 &
  23.69 \\
 &
  \checkmark &
  \checkmark &
  \checkmark &
   &
  30.89 &
  38.08 &
  37.59 &
  30.10 &
  23.59 &
  16.90 &
  15.50 &
  31.98 &
  33.91 &
  33.41 &
  30.93 &
  27.85 \\
 &
  \checkmark &
  \checkmark &
   &
  \checkmark &
  31.24 &
  37.96 &
  38.67 &
  30.21 &
  24.37 &
  17.81 &
  15.89 &
  32.86 &
  33.75 &
  \textbf{33.75} &
  31.09 &
  28.31 \\
 &
  \checkmark &
  \checkmark &
  \checkmark &
  \checkmark &
  \textbf{31.30} &
  \textbf{38.15} &
  \textbf{38.66} &
  \textbf{30.96} &
  \textbf{24.73} &
  \textbf{18.60} &
  \textbf{16.12} &
  \textbf{33.14} &
  \textbf{33.97} &
  31.48 &
  \textbf{31.30} &
  \textbf{28.43} \\ \hline
\end{tabular}}
\end{table*}


\textbf{Ablation Study}. In Tab.\ref{tab:abla}, we discuss the effect of each part in MID including meta framework (i.e., Meta), teacher-student distillation (i.e., DST) and the multi-consistency constraints (i.e., AC, CC) under white box attack. Note that without Meta and DST, MID degrades into an ERM-like model \cite{vapnik1999overview}. Without DST, MID becomes a meta-AT framework. From Tab.\ref{tab:abla}, each component is crucial for improving robustness. The multi-consistency constraints also provide significant gains against each attack. The best performance is always achieved when all components work together. 

\textbf{Evaluation on Larger Backbones of Teacher Model}. Considering that some defensive distillation methods also use larger teacher models than student models \cite{zi2021revisiting}, we discuss the impact of larger teacher models in MID based on CIFAR10 dataset. As presented in Tab.\ref{tab:differ_teacher}, the larger teacher models bring better adversarial robustness to known attacks, 
but do not show clear gain to unknown attacks. An explanation is that, the distillation from a larger teacher is too tight and stiff, which may not perform ideal transfer between larger teacher and smaller student. However, the alignment between teacher and students of the same structure is easier, and the teacher model is more likely to transfer student model \textit{knowledge} rather than \textit{answers} (i.e., learning invariance). An analogy is that it may be difficult to have a college professor teach the primary school pupils, but it is easier for peer students to discuss with each other.

\begin{table*}[]
\caption{Results of larger teacher models on CIFAR10. For the student encoder evaluation, we still use the vanilla settings.}
\centering
\scriptsize
\setlength{\tabcolsep}{2.5mm}{
\label{tab:differ_teacher}
\begin{tabular}{c|cccc|ccccccc}
\hline
\multirow{2}{*}{Teacher backbone} & \multicolumn{4}{c|}{Known Attack}      & \multicolumn{7}{c}{Unkonwn Attack}                             \\ \cline{2-12}
                         & PGD$\rm _N$  & PGD$\rm _T$           & MIM$\rm _N$  & MIM$\rm _T$  & FGSM$\rm _N$ & FGSM$\rm _T$ & BIM$\rm _N$  & BIM$\rm _T$  & CW$\rm _N$   & CW$\rm _T$            & AA$\rm _N$   \\ \hline
ResNet18 (base) &
  69.88 &
  66.47 &
  65.08 &
  45.12 &
  41.63 &
  \textbf{45.09} &
  62.24 &
  \textbf{59.39} &
  64.42 &
  64.48 &
  \textbf{53.29} \\
ResNet50                 & 70.06 & \textbf{74.13} & 69.16 & 51.53 & 41.59 & 42.93 & 66.53 & 58.20 & 65.29 & 65.27          & 46.98 \\
ResNet101 &
  \textbf{71.23} &
  73.33 &
  \textbf{71.03} &
  57.97 &
  \textbf{44.71} &
  44.70 &
  \textbf{67.81} &
  56.95 &
  62.97 &
  64.70 &
  50.20 \\
DenseNet201 &
  70.30 &
  72.31 &
  70.96 &
  \textbf{59.58} &
  43.59 &
  44.38 &
  62.31 &
  55.11 &
  \textbf{65.52} &
  65.44 &
  49.03 \\
ConvNext-Large           & 70.19 & 69.14          & 70.95 & 57.16 & 44.54 & 45.00 & 62.00 & 58.11 & 61.71 & \textbf{65.94} & 51.08 \\ \hline
\end{tabular}}
\end{table*}

\begin{figure*}
\centering
\includegraphics[height=5cm]{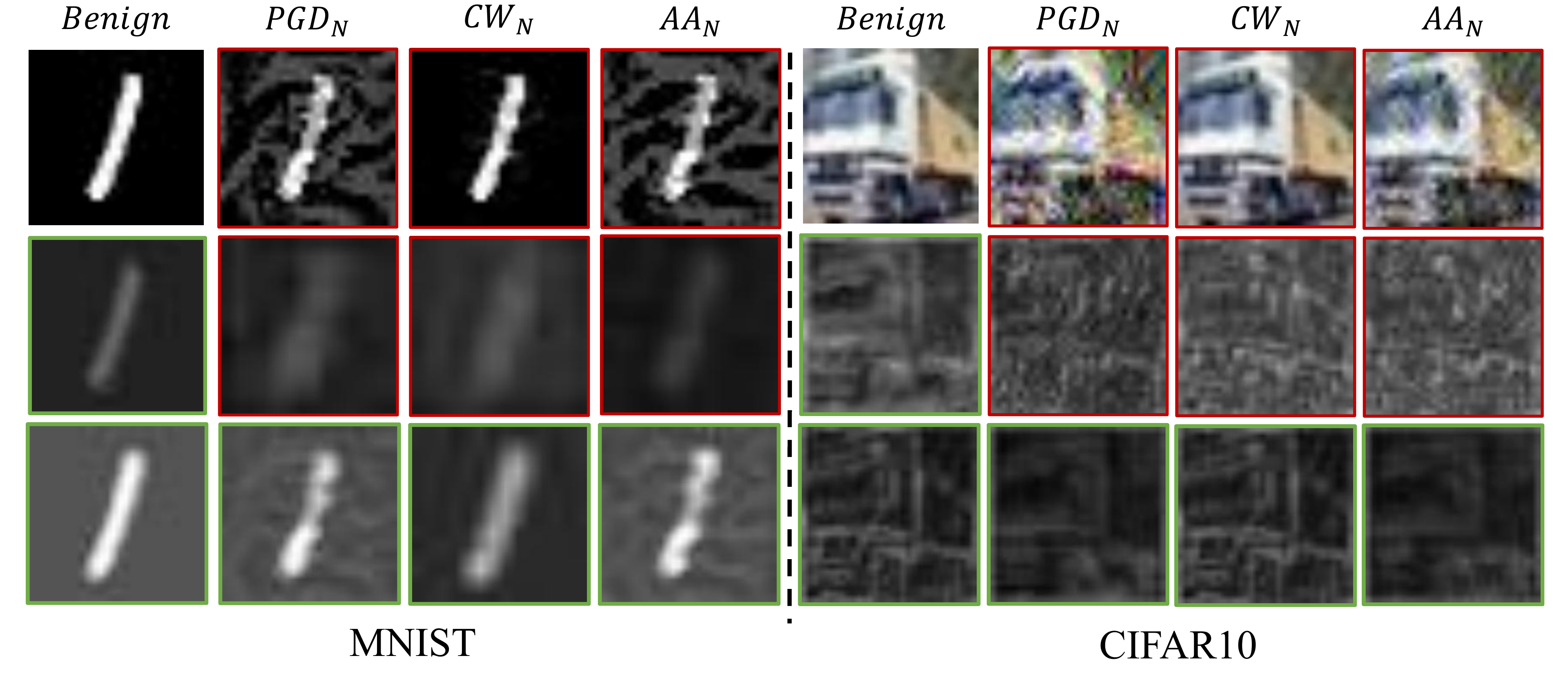}
\caption{Gradient visualization of MNIST and CIFAR10. The $1^{st}$ row denotes the benign samples and adversarial samples with different attacks. The $2^{nd}$ and $3^{rd}$ row denotes the gradients of the baseline and MID, respectively. Obviously, the gradients from MID show clear semantics. 
}
\label{fig:feat}
\end{figure*}

Besides that, we also have two interesting observations:

$\bullet$ \textbf{Easy to attack does not mean hard to defend, and vice versa.} By comparing Tab.\ref{tab:1} with Tab.\ref{tab:2}, although the attack power on the baseline of the black-box attack  (e.g., 10.33\% of $\rm MIM_N$ on CIFAR100) is obviously inferior to white-box attack (e.g., 0.94\% of $\rm MIM_N$ on CIFAR100), the robustness of MID to black-box attack (e.g., 24.41\% of $\rm MIM_N$ on CIFAR100) is often weaker than white-box attack (e.g., 30.96\% of $\rm MIM_N$ on CIFAR100). This suggests that attacks cannot be simply described as `strong' or `weak' but is related to the preference of the model. Black box attack is a little weaker than white box attack since the attacker cannot fully understand the target model. So a possible reason is, defenders cannot fully learn the attack strategy and model basis of the black box attackers, thus resulting in robustness degradation than white-box attacks. Additionally, the attacks used in the defense can be considered as white-box attacks, so there lacks the prior of black-box attacks to some extent, which may explain why robustness to white-box attacks is often better. \cite{yuan2021meta} learns the invariance among various models to attack, which may improve the robustness to black-box attacks since it queries the distribution of multiple models.

$\bullet$ \textbf{Accuracy degradation of benign samples is improved, but still a challenge.} The accuracy of MID for benign samples always decreases slightly, which is a common problem for defense methods based on Adversarial Training. The lower the original accuracy is, the greater the decrease is (e.g., CIFAR100 in our experiments). We think it is exactly due to the model begins to pay attention to the attack invariance information between benign and adversarial samples, such that the model ignores some unstable details and thus loses some accuracy. The analysis on the interpretability of MID in next section also confirms our conjecture. So far, the accuracy degradation of the benign samples is still a challenge for adversarial training and its variants.

\subsection{Interpretability of MID} \label{sec4_3}
Robust model should focus on the attack-invariance features and  have strong intrinsic interpretability. Thus, 
we analyze MID from four aspects: 1) gradient interpretability, 2) feature distribution, 3) structural preference, and 4) attention map. 

\textbf{(1) Gradient interpretability with rich semantics.} The gradient here refers to the gradient of the loss function w.r.t. the sample, whose essence is the weight assigned by the model to each pixel of the sample.
Both \cite{ross2018improving} and \cite{wang2021demystifying} claim that the gradients and features of the robust model are more informative in semantics. We randomly select some images from CIFAR10 and MNIST, and visualize the gradient of each sample, as shown in Fig.\ref{fig:feat}. Obviously, the gradient of the baseline is chaotic, while more interpretable semantic information appears in the gradient of MID, which 
well evaluates the robustness of MID since richer semantics in gradients mean the network is more sensitive to semantic features rather than adversarial perturbations. Also, models with rich semantics in their gradient can perform better to resist the gradient based perturbation, since its gradient is harmonious with the semantics features. This further confirms the robustness of MID to different attacks, since similar semantic in gradients is remained for all attacks. 


\begin{figure*}
\centering
\includegraphics[height=5.5cm, width=15cm]{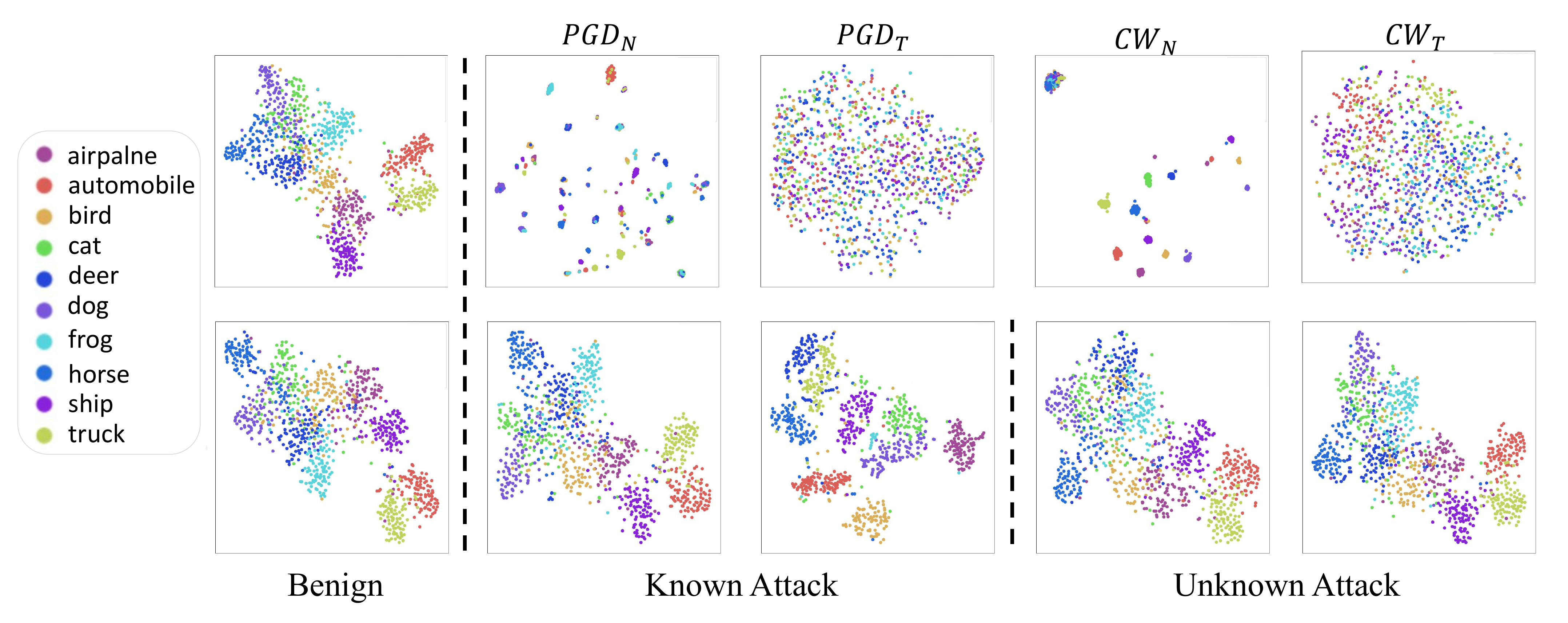}
\caption{T-SNE diagram of features encoded by baseline (the $1^{st}$ row) and MID (the $2^{nd}$ row) on CIFAR10. To ensure readability, we screen part of the known attacks and unknown attacks. Intuitively, the adversarial feature distribution of Baseline is chaotic while that of MID is more recognizable.}
\label{fig:tsne}
\end{figure*}

\begin{figure*}
\centering
\includegraphics[height=4cm]{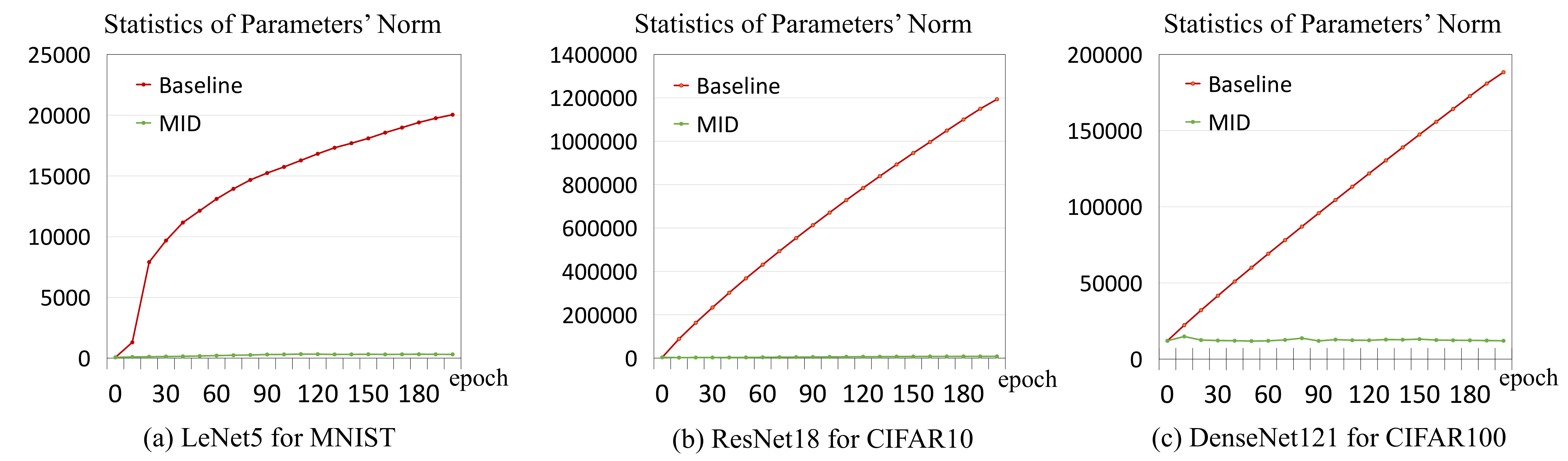}
\caption{Variation curve of the models' sparsity in the training process. Sparsity index of model is defined as $\mathrm{I}_{sparse }=\sum\|\theta\|^{2}$. 
Baseline seems to endlessly fit the training set with more parameters while MID learns more stable invariant features through sparser parameters and weights.}
\label{fig:para}
\end{figure*}

\begin{figure*}
\centering
\includegraphics[height=3.8cm, width=15cm]{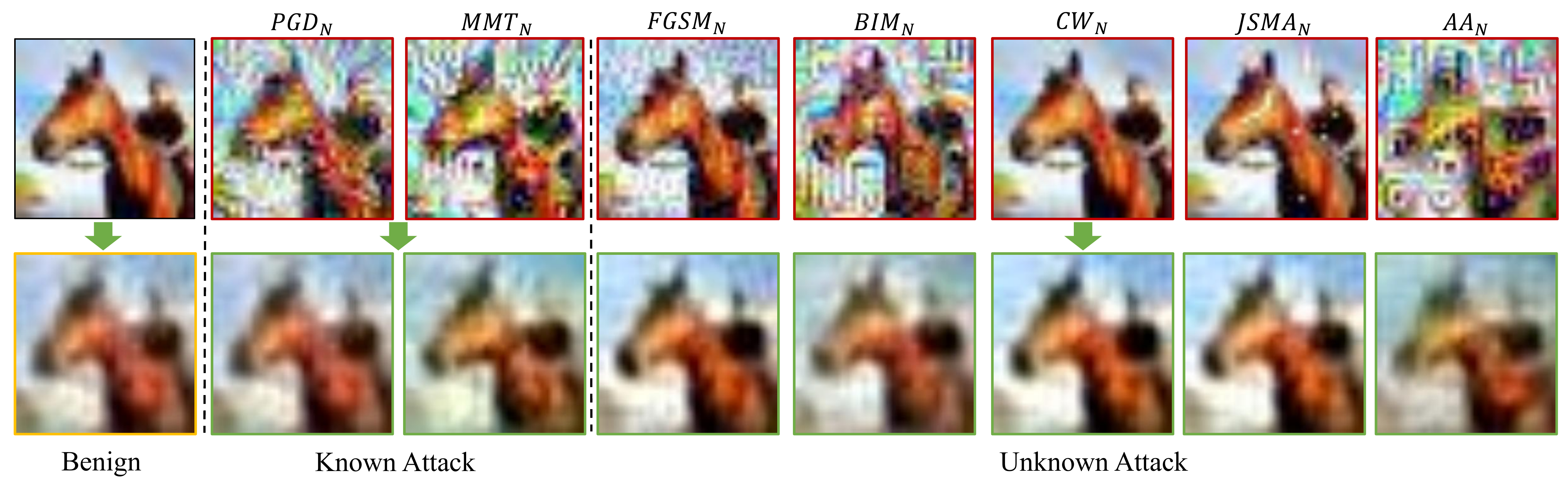}
\caption{The feature decoder $D(\theta_{D})$ of teacher module is used to decode the features encoded by student encoder $E_{student}(\theta_{E}^{s})$. The $1^{st}$ row denotes part of the adversarial samples, and the $2^{nd}$ row shows the decoded samples via MID. 
}
\label{fig:decoded}
\end{figure*}

\textit{Why do semantic features appear in MID?} \cite{wang2021demystifying} argues that adversarial training not only focuses data distribution, but also implicitly explores the distribution of model parameters, so it has some ability to extrapolate and generate some pictures with semantics. Thus, the gradient of the MID model contains more interpretable information. In fact, we have analyzed that the meta learning framework in MID explicitly finds and retains parameter groups with similar activations for various attacks, so the joint distribution of model parameters and data is explored. From the perspective of attack principle, the rich semantic information in the gradient further weakens the possibility being attacked. For the gradient-based attack, an intuitive understanding is when the noise generated from gradient is used as attack, it may have a greater chance to disturb the benign features. 

\textbf{(2) MID learns more discernible feature distributions.} We have analyzed MID from the perspective of manifold learning in Sec.\ref{feasibility}. To further analyze the interpretability of MID and verify our conjecture about manifolds, we use T-distributed Stochastic Neighbor Embedding (T-SNE) to visualize the feature distribution of MID and baseline models on CIFAR10, as shown in Fig.\ref{fig:tsne}. The figures in the first and second row are the feature visualizations of baseline and MID, respectively. All features are sampled from the high-level features from encoder used for classification.

Intuitively, the feature distribution of benign samples encoded by baseline is interpretable. The features of the same class are clustered and separated from other classes. However, the feature distribution of adversarial samples encoded by baseline is chaotic, and the features under non-targeted attack seem to be randomly distributed in feature space, or crowded in a narrow neighborhood under targeted attack. Obviously, the classifier of the baseline can not effectively recognize the features of such distribution under attacks. On the contrary, MID restores the cluster-like feature distribution and provides more recognizable features even under different attacks. This verifies our conjecture that MID can restore the feature manifold and the features are attack-invariant, semantically-related and recognizable.

\textbf{(3) MID is sparser and benign-preservable in model parameters.} We have analyzed that, to learn invariant features among various attacks and filter out unstable features,  MID will select stable parameters with common activations for all attacks and discard those unstable parameters. Then, MID becomes much sparser than the baseline under the same backbone. Therefore, we compute the model sparsity index of the baseline and MID within 200 epochs as shown in Fig.\ref{fig:para}. The results show that the parameter norm of the baseline model is increasing and obviously much larger than that of MID. Additionally, no matter what backbone and dataset is used (e.g., LeNet5, ResNet18, ResNet 121 on MNIST, CIFAR10 and CIFAR100), the model complexity of MID only increases slightly in the first 10 epochs, and then remains stable. This verifies our conjecture about the model sparsity of MID that it only focuses on parameters with common activations to all attacks in the hypothesis space.

It is worth mentioning that our analysis of model sparsity does not negate the  gain brought by the depth of the model, nor does it suggest that the MID should directly adopt a simpler model. On the one hand, a deeper model provides a broader hypothesis space for the meta learning framework, while the representation ability of the simple model may be insufficient. For example, for a polynomial regression task, we hope that our hypothesis space contains as many higher-order terms as possible, although some of them may not work. Effective regularization methods can filter these useless high-order terms. On the other hand, if the high-order terms in the hypothesis space are insufficient, the task may easily exceed the upper bound of the representation ability of the proposed model. Our analysis indicates that MID has a strong regularization ability.

To explore the intuitive interpretability of the features learned by MID, we use the decoder $D(\theta_D)$ of the teacher model to decode the features, and the regenerated images are shown in Fig.\ref{fig:decoded}, where the first row shows benign samples and adversarial samples under different attacks and the second row shows regenerated samples. First of all, all the regenerated pictures no matter from which kind of adversarial sample are highly similar. This means the student encoder $E_{student}$ has captured similar but benign-preservable features from all attacks, which is exactly the attack-invariant information between various adversarial samples and benign samples. Although a small amount of detailed information is lost, the regenerated image contains almost no adversarial perturbation and preserves the semantics of benign images. In other words, MID filters out the attack-related features from various attacks and retains features only relating to semantics. In this sense, the student encoder of MID self-contains the function of denoising and regeneration, but it is not the focus of this work.

Further, from Fig.\ref{fig:decoded}, the regenerated image (robust sample) looks more like a low-frequency component of a benign sample, which also suggests that the low-frequency information may be the attack-invariant robust information.  We will discuss this topic in the next section. The above qualitative analyses indicate that the baseline overfits the unstable features with redundant parameters, while MID learns attack-invariant features with sparser parameters.

Notably we show the regenerated images only to analyze the feature bias of MID, but the essence of MID is not a real denoiser. In fact, directly feeding the regenerated images into the classification network may perform badly, 
due to the decoder's bias. Therefore, it may be interesting to extend the MID towards a plug-and-play extrapolator.

\begin{figure}
\centering
\includegraphics[height=3.4cm, width=8.0cm]{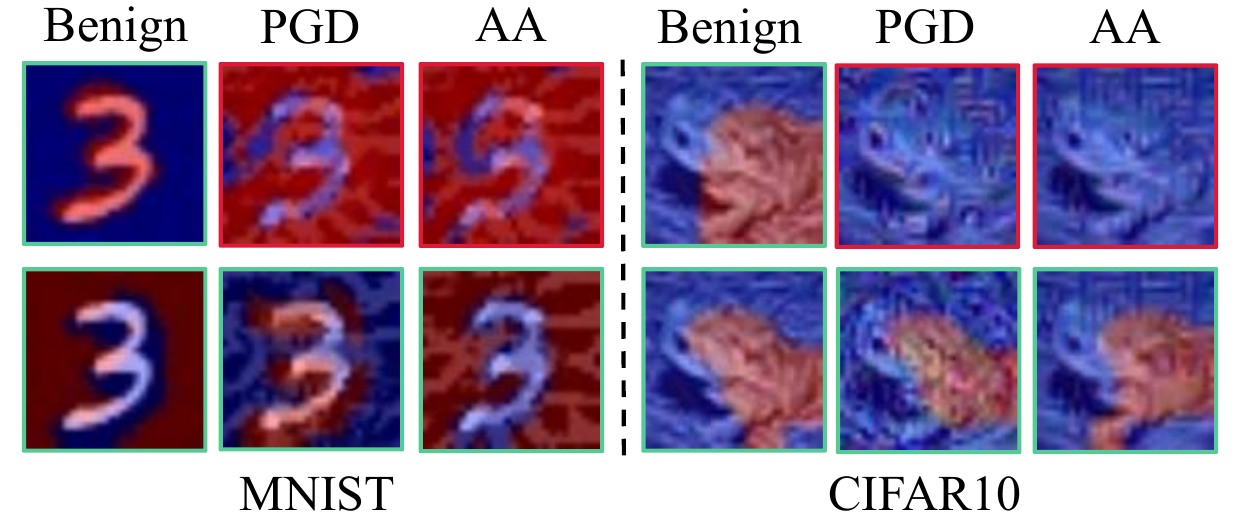}
\caption{Attention maps of the baseline (the 1st row) and MID (the 2nd row). $PGD_N$ is known attack and $AA_N$ is unknown attack. Intuitively, MID is semantic-focused and more interpretable than baseline model.}
\label{fig:CAM}
\vspace{-0.3cm}
\end{figure}

\textbf{(4) Attention map of MID focuses on semantics.}
Attention map reflects the regional preference of DNNs and provides an intuitive interpretability on adversarial robustness. The attention map of a robust model has rich semantics. 

As shown in Fig.\ref{fig:CAM}, the attention map of the baseline w.r.t. the benign sample focuses on the semantic part. For example, the outline of the frog's back is a significant semantic part for both baseline and the robust model. Therefore, it has obvious semantic contour, especially in MNIST, where the boundary between semantics and background is clear and the attention map almost perfectly separates the digits from background. However, the attention maps of the baseline w.r.t. the adversarial samples are chaotic and irregular, which explains the vulnerability of DNN (i.e., baseline model). In contrast, MID obtains interpretable attention maps of adversarial samples, similar to the attention map of the baseline w.r.t. the benign samples. For example, MID pays attention to the frog's body shape and outline of digits, and shows adversarial robustness against known and unknown attacks. An interesting phenomenon is that in MNIST, the robust attention map may reverse and focus on the background to some extent. However, it does not affect the semantic discrimination and adversarial robustness.

\section{Discussion}
\subsection{Is LFC the attack-invariant feature?}
\begin{figure*}
\centering
\includegraphics[height=4cm, width=17cm]{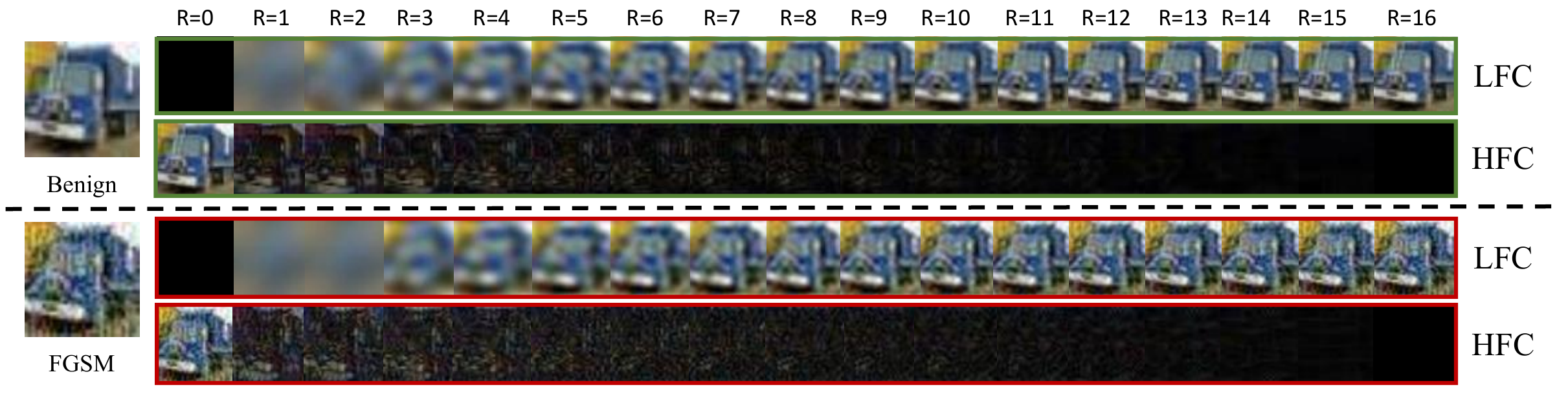}
\caption{Low- and high-pass filters are performed on benign samples and adversarial samples (FGSM), in order to obtain and visualize the Low- and High-Frequency components (LFC vs. HFC) respectively. $R$ refers to the cut-off frequency of the filter.}
\label{fig:disfig}
\end{figure*}

\begin{figure*}
\centering
\includegraphics[height=3.8cm, width=16.5cm]{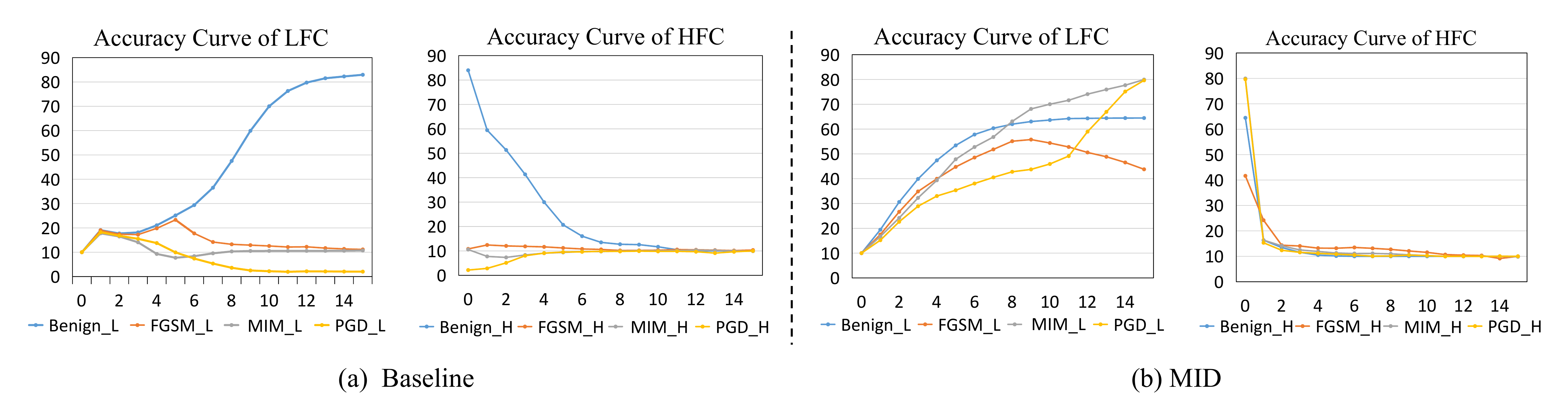}
\caption{Accuracy curve of Baseline (a) and MID (b) to high- and low-frequency components with respect to the increasing cut-off frequency for benign samples and adversarial samples under different attacks.}
\label{fig:distable}
\end{figure*}

The regenerated samples in Fig.\ref{fig:decoded} suggest that the low-frequency components (LFC) may be the attack-invariant features among various adversarial samples. Therefore, we briefly discuss this issue in this section. We transform the benign and adversarial samples into the frequency domain by \emph{Fast Fourier Transform (FFT)}, and feed them into high- and low-pass filters respectively. Then inverse FFT is conducted to obtain high-frequency components (HFC) and low-frequency components samples. Generally, we change the cut-off frequency $R$ of the filters from 0 to 16, since the image size is [3, 32, 32]. The filtered image is shown in Fig.\ref{fig:disfig}. Further, we exploit the baseline and MID model to predict the obtained high- and low-frequency components respectively. The variation curve of the accuracy w.r.t. the increasing cut-off frequency $R$ is shown in Fig.\ref{fig:distable}. We have the following observations:

$\bullet$ Visually, the LFC of benign and adversarial samples look similar. When $R<7$, it is almost difficult to capture the noise in the low-frequency part of the adversarial samples, and the accuracy of LFC basically increases linearly.

$\bullet$ For the baseline, when $R<5$, the accuracy of LFC for benign and adversarial samples is almost the same. However, when $R>5$, the accuracy of adversarial samples is plummeted. It shows that the low-frequency components of benign and adversarial samples are similar, and the perturbations mostly exist in the high-frequency part. In contrast, the rising trend of MID for benign and adversarial samples is basically the same, which proves that MID prefers to learning attack-invariant low-frequency information.

$\bullet$ For the high-frequency component of benign samples, the baseline performs well, but is not sensitive to that of the adversarial sample, which further shows that the adversarial noise mainly exists in the high-frequency part. The declining trend of MID w.r.t. high-frequency components is steeper, which also confirms that adversarial robust semantics mainly exist in LFCs, but MID is capable to learn low-frequency information. Naturally, MID can be more robust.

In short, for benign samples, the accuracy of the baseline model rises with the increasing abundance of the high-frequency information. However, for the adversarial samples, with the influx of high-frequency perturbations, the accuracy of the baseline for the adversarial samples is deteriorated. On the other hand, the accuracy of MID for the spectral components always rise with the increasing abundance of the spectrum information, no matter for benign or adversarial samples. This is because MID is more robust to the HFCs with adversarial perturbations.

In fact, the above observations are coincidentally with the perspectives of \cite{wang2020high,  huang2022adversarially} that the model can learn basic recognition ability from the low-frequency component (LFC), and then learn the fine-grained information from the high-frequency component (HFC). That is, the deep model will first learn some generalizable features (e.g., the overall shape of the animals) and then learn the fine-grained features associated with the task (e.g., animals' hair, facial features, etc.). In fact, the research on GAN and AutoEncoder in the community can easily prove this claim, because the generated model always generates some rough color patches first, and then the details are filled in.

Although HFC is more discernible, it lacks commonality and invariance among various attacks. For example, for video surveillance, the police can conduct preliminary investigation of suspects according to their body shape (LFC). The detailed information (HFC) such as clothing can help the police further determine the identity of the suspect. However, the clothing information is not robust, because the suspect can change his/her clothes to escape. From this perspective, the models focused on low-frequency components may be more robust and generalizable, but inaccurate. On the contrary, the models focused on high-frequency components may be more accurate, but vulnerable and unstable. \cite{jia2022exploring, huang2022adversarially, luo2022frequency} propose that the adversarial perturbation playing a major role in adversarial attack mainly exists in the HFCs, so the attack-invariant robust feature among various adversarial samples mainly exists in the LFC. Our analysis also provides an explanation and a direction for the trade-off between accuracy and adversarial robustness.

Despite the above empirical observations, the mathematical boundary between the semantic features and adversarial noise in frequency domain is difficult to be given accurately, which, therefore, 
is still an open topic.

\subsection{Cross-validation Training of MID}
Cross-validation (CV) is a commonly used training protocol to evaluate the generalization ability. A task close to our work is Domain Generalization \cite{kang2022style}, where researchers usually employ leave-one-domain-out strategy for cross-validation because the public dataset has a very few domains. In our main setting, we do not adopt CV strategy because the number of attacks is too large and it will cost unaffordable training time via leaving-one-attack-out strategy. Additionally, generating adversarial samples for each attack is also time consuming. Therefore, we fix the known attacks (domains) during training as the main setting to pursuit an attack-invariant model transferrable to unknown attacks.
To verify the effectiveness of MID under the CV protocol, we perform cross-validation experiments by selecting four commonly used attacks, i.e., PGD, MIM, BIM and FGSM, in which 3 attacks are used for training and the remaining one is used for test. The experimental results are shown in Tab. \ref{tab:cross}, from which we see MID achieves excellent adversarial robustness in various settings. Nevertheless, we may not advocate the cross-validation setting 
for adversarial defense task due to very expensive training cost, and fixing the attacker pool is suggested. Overall, in each setting, the superiority of MID is fully evaluated.


\begin{table}[t]
\centering
\scriptsize
\caption{Robustness evaluation under cross-validation protocol.}
\label{tab:cross}
\begin{tabular}{c|c|ccc}
\hline
\multicolumn{1}{c|}{Datasets} & \multicolumn{1}{c|}{Test attack}                                               & \multicolumn{1}{c|}{Method}                  & Accuracy       & Robustness     \\ \hline
\multirow{8}{*}{MNIST}   & \multirow{2}{*}{\begin{tabular}[c]{@{}l@{}} PGD$_{\rm N}$ \end{tabular}} & \multicolumn{1}{c|}{AT} & 98.70 & 97.23 \\
                             &                                                                           & \multicolumn{1}{c|}{MID} & \textbf{99.18} & \textbf{99.14} \\ \cline{2-5}
                             & \multirow{2}{*}{\begin{tabular}[c]{@{}l@{}} MIM$_{\rm N}$ \end{tabular}} & \multicolumn{1}{c|}{AT}  & 98.70          & 96.47          \\
                             &                                                                           & \multicolumn{1}{c|}{MID} & \textbf{99.16} & \textbf{99.13} \\ \cline{2-5}
                             & \multirow{2}{*}{\begin{tabular}[c]{@{}l@{}} BIM$_{\rm N}$ \end{tabular}} & \multicolumn{1}{c|}{AT}  & 98.70          & 93.94          \\
                             &                                                                           & \multicolumn{1}{c|}{MID} & \textbf{99.23} & \textbf{99.19} \\ \cline{2-5}
                             & \multirow{2}{*}{\begin{tabular}[c]{@{}l@{}} FGSM$_{\rm N}$ \end{tabular}} & \multicolumn{1}{c|}{AT}  & 98.70          & 90.99          \\
                             &                                                                           & \multicolumn{1}{c|}{MID} & \textbf{99.08} & \textbf{95.46} \\ \hline
\multirow{8}{*}{CIFAR10} & \multirow{2}{*}{\begin{tabular}[c]{@{}l@{}} PGD$_{\rm N}$ \end{tabular}} & \multicolumn{1}{c|}{AT} & \textbf{63.24} & 57.70 \\
                             &                                                                           & \multicolumn{1}{c|}{MID} & 60.46          & \textbf{59.02} \\ \cline{2-5}
                             & \multirow{2}{*}{\begin{tabular}[c]{@{}l@{}} MIM$_{\rm N}$ \end{tabular}} & \multicolumn{1}{c|}{AT}  & \textbf{63.24}          & 57.45          \\
                             &                                                                           & \multicolumn{1}{c|}{MID} & 59.16          & \textbf{60.72} \\ \cline{2-5}
                             & \multirow{2}{*}{\begin{tabular}[c]{@{}l@{}} BIM$_{\rm N}$ \end{tabular}} & \multicolumn{1}{c|}{AT}  & \textbf{63.24}          & 55.77          \\
                             &                                                                           & \multicolumn{1}{c|}{MID} & 60.98          & \textbf{60.95} \\ \cline{2-5}
                             & \multirow{2}{*}{\begin{tabular}[c]{@{}l@{}} FGSM$_{\rm N}$ \end{tabular}} & \multicolumn{1}{c|}{AT}  & \textbf{63.24}         & 35.72          \\
                             &                                                                           & \multicolumn{1}{c|}{MID} & 58.11          & \textbf{69.56} \\ \hline
\end{tabular}
\end{table}

\begin{table}[t]
\centering
\scriptsize
\caption{Performance trained on larger Attacker Pool.}
\label{tab:more}
\begin{tabular}{l|c|ccccc}
\hline
Datasets                 & Attacker Pool   & benign & AA    & DDN   & FAB   & stAdv \\ \hline
\multirow{3}{*}{MNIST}   & Basic       & 98.83  & 98.24 & 95.47 & 95.18 & 91.22 \\
                         & +FGSM  & 98.52  & 99.02 & 98.49 & 98.49 & 93.31 \\
                         & +FGSM$\&$BIM & 98.13  & 99.04 & 98.93 & 98.89 & 93.80 \\ \hline
\multirow{3}{*}{CIFAR10} & Basic       & 64.48  & 53.29 & 54.71 & 55.37 & 64.61 \\
                         & +FGSM  & 60.71  & 53.88 & 58.58 & 59.40 & 67.10 \\
                         & +FGSM$\&$BIM & 58.42  & 54.00 & 58.46 & 59.41 & 66.91  \\ \hline
\end{tabular}
\end{table}


\subsection{Training on Larger Attacker Pool}
The proposed MID aims to achieve generalizable adversarial robustness to unknown attacks, which, obviously, is a tricky but under-studied challenge in the community. In MID, totally 4 kinds of attacks consisting of the targeted and non-targeted versions of PGD and MIM are included in the basic \emph{Attacker Pool}, which can be enlarged freely. Intuitively, it is rational to improve the generalization by enlarging the \textit{Attacker Pool}. Therefore, we conduct experimental verification on this perspective. We consider two new settings: +FGSM and +FGSM$\&$BIM on the basic attacker pool. The former means FGSM attacker is added in the \emph{Attacker Pool} and the latter means both FGSM and BIM attackers are added. Note that the choice of the pool is independent of the training protocol. The results of MID under different \textit{Attacker Pool} are presented in Tab.\ref{tab:more}. Clearly, the generalization performance can be improved by enlarging the attacker pool, which complies with our conjecture. Inevitably, the training cost becomes larger and discussed in next section.

\subsection{Limitations and Failures of MID}\label{sec5_2}

\textbf{Accuracy vs. Robustness.} As can be seen from Tab.\ref{tab:1} and Tab.\ref{tab:2}, despite achieving generalizable adversarial robustness and average robustness, similar to \cite{zhang2019theoretically}, MID does not fully address the trade-off between accuracy (benign samples) and robustness (adversarial samples), a popular but tricky problem of adversarial training. That is, accuracy degradation on benign data is encountered. This may be caused by data distribution shift and model capacity. 
Interestingly, the experiments on ImageNet preliminarily verify that the training paradigm of 'large model + fine-tuning' may alleviate this issue, i.e., MID achieves almost similar accuracy with original model (79.56\% vs. 81.91\%).

\textbf{Computation Cost.} As discussed in Tab.\ref{tab:more}, a larger Attacker Pool can help learn better attack-invariance, which, however, causes an increase in computational cost. We present the training time per epoch on different datasets in Tab.\ref{tab:time}, and MID takes 5 times longer time than AT. We observe that MID is comparable to ERM \cite{vapnik1999overview} in computation cost, but shows better performance than ERM as shown in Table~\ref{tab:abla} (the first and last rows for each dataset). Further, we use WRN-34-10 backbone for AT training which is 5 times larger than our used Resnet-18 under white-box setting, and the average performance on CIFAR10 is 55.95\%, lower than our 59.25\% (see Table~\ref{tab:1}). Thus, merit of MID is indicated.

\begin{table}[]
\centering
\scriptsize
\caption{Comparison of training time per epoch for vanilla AT, ERM, and MID. Notably MID is trained with PGD and PGD$\&$MIM, respectively.}
\label{tab:time}
\setlength{\tabcolsep}{3mm}{
\begin{tabular}{c|ccc}
\hline
\multirow{2}{*}{Defense model} & \multicolumn{3}{c}{Time cost (s)} \\ \cline{2-4}
                        & MNIST  & CIFAR10  & CIFAR100  \\ \hline
Baseline                & 4$\pm$1    & 18$\pm$2     & 4$\pm$1       \\
Madry's AT              & 45$\pm$5   & 590$\pm$10   & 960$\pm$20    \\
ERM                     & 160$\pm$5  & 2300$\pm$10  & 3200$\pm$20   \\
MID$_{\rm PGD}$                  & 100$\pm$5  & 1350$\pm$10  & 1870$\pm$20   \\
MID$_{\rm PGD\&MIM}$                 & 180$\pm$5  & 2530$\pm$10  & 3460$\pm$20   \\ \hline
\end{tabular}}
\end{table}

\begin{table}[]
\centering
\scriptsize
\caption{Performance trained on the attacker pool without targeted attackers.}
\label{tab:veri}
\setlength{\tabcolsep}{1mm}{
\begin{tabular}{c|cccccc}
\hline
Test attack        & $\rm PGD_T$ & $\rm MIM_T$ & $\rm BIM_T$ & $\rm FGSM_T$ & $\rm CW_T$ & $\rm JSMA_T$ \\ \hline
MNIST    & 98.35  & 94.11  & 97.46  & 90.45   & 98.74 & 41.33   \\
CIFAR10  & 25.01  & 20.99  & 31.68  & 26.57   & 42.09 & 47.35   \\
CIFAR100 & 24.56  & 13.08  & 32.12  & 6.79    & 31.96 & 31.02   \\ \hline
\end{tabular}}
\end{table}

\textbf{Insufficient Generalized Robustness.} Despite taking a step forward on the generalization of robustness, the performance of unknown attacks is related to the number of known attacks in the attacker pool. 
When there are gradient-based (MIM) attacks in the Attacker Pool, MID is more generalizable to gradient-based attackers (e.g., FGSM). Similarly, when the model suffers from unknown targeted attacks (e.g., $BIM_T$), training on an Attacker Pool containing targeted attacks (e.g., $PGD_T$) is helpful. This empirically verifies when there lacks targeted attack in the Attacker Pool, the robustness of MID to unknown target attacks is greatly degraded, by comparing Tab.\ref{tab:veri} with Tab.\ref{tab:1} and Tab.\ref{tab:2}. Notably, in Tab.\ref{tab:veri}, we only use non-targeted attacks $PGD_N$, $MIM_N$ and the benign samples to train MID, and the targeted attacks for test. However, the fact is we cannot access full prior knowledge of all unknown attacks, which, therefore, is still a challenge of domain generalization (DG). We may suggest further enhance the generalization of MID by stochastically generating the attacker pool in order to find the attack invariance. Moreover, the current popular test-time adaptation (TTA) training paradigm is also suggested by adapting new test attackers on-line. 


\section{Conclusion and Future Work}

To achieve the generalizable robustness of adversarial defense against unknown attacks, we propose a new attack-invariant defense method based on meta learning, called Meta Invariance Defense (MID). By iteratively performing the defense against (simulated) known attacks and adapting the defense model against (simulated) unknown attacks, the proposed MID network progressively emphasizes the parameters having stable activations for all attacks. Then a multi-consistency protocol composed of label consistency, adversarial consistency and cyclic consistency is further proposed to facilitate the learning of attack-invariant robust features among various adversarial samples from pixel, feature and prediction levels, towards the universal robustness. Theoretical and experimental analysis illustrate the rationality and effectiveness of the proposed MID approach. In-depth insights on the interpretability and spectral components are presented.

This paper explores the possibility of pursuing generalizable adversarial robustness to future attacks by learning attack-invariance, which complies with the universal scenarios of domain generalization (DG). This has been imposed as a big challenge in machine learning and computer vision community. Although a great progress of DG has been made towards the domain invariance, few approaches really work in real-world scenarios by frozen the model trained off-line and becomes frustrated. Especially, the types of future adversarial attacks are endless. Therefore, inspired by the nature of big models that greatly benefit the down-stream tasks, \textit{test-time training regimes} or \textit{on-line adaptation} of new attacks in the future can be vividly portrayed.


%

\appendices


%
%

\ifCLASSOPTIONcaptionsoff
  \newpage
\fi



%

\bibliographystyle{IEEEtrans}
\bibliography{IEEEexample}
%

\begin{IEEEbiography}[{\includegraphics[width=1.2in,height=1.2in,clip,keepaspectratio]{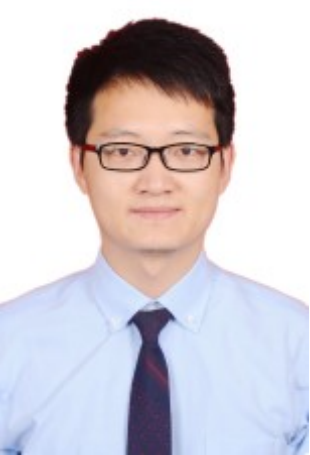}}]{Lei Zhang}
(M'14-SM'18) received his Ph.D degree in Circuits and Systems from the College of Communication Engineering, Chongqing University, Chongqing, China, in 2013. He worked as a Post-Doctoral Fellow with The Hong Kong Polytechnic University, Hong Kong, from 2013 to 2015. He is currently a Professor with Chongqing University. He has authored more than 100 scientific papers in top journals and conferences, including IEEE Transactions (e.g., T-PAMI, T-IP, T-MM, T-CSVT, T-NNLS), CVPR, ICCV, ECCV, ACM MM, AAAI, IJCAI, etc. He is on the Editorial Boards of several journals, such as IEEE Transactions on Instrumentation and Measurement, Neural Networks (Elsevier), etc. Dr. Zhang was a recipient of the 2019 ACM SIGAI Rising Star Award. His current research interests include deep learning, transfer learning, domain adaptation and computer vision. He is a Senior Member of IEEE.
\end{IEEEbiography}

\begin{IEEEbiography}[{\includegraphics[width=1.2in,height=1.2in,clip,keepaspectratio]{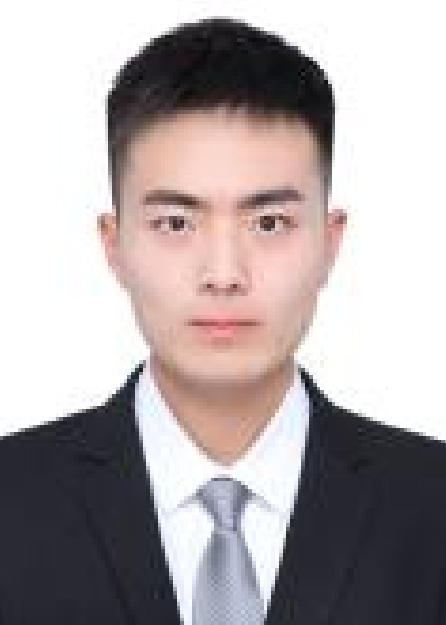}}]{Yuhang Zhou}
Yuhang Zhou received the B.S. degree from the Shandong University, China, in 2020. He is currently persuing the M.S. degree at Chongqing University, China.

His research interests include deep learning, computer vision, and adversarial robustness.
\end{IEEEbiography}

\begin{IEEEbiography}[{\includegraphics[width=1in,height=1.25in,clip,keepaspectratio]{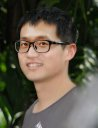}}]{Yi Yang}
received the Ph.D. degree in computer science from Zhejiang University, China, in 2010. He was a Post-Doctoral Research with the School of Computer Science, Carnegie Mellon University, USA. He is currently a Professor with the University of Technology Sydney, Australia. His current research interest includes machine learning and its applications to multimedia content analysis and computer vision, such as multimedia indexing and retrieval, surveillance video analysis, and video semantics understanding.
\end{IEEEbiography}

\begin{IEEEbiography}[{\includegraphics[width=1in,height=1.25in,clip,keepaspectratio]{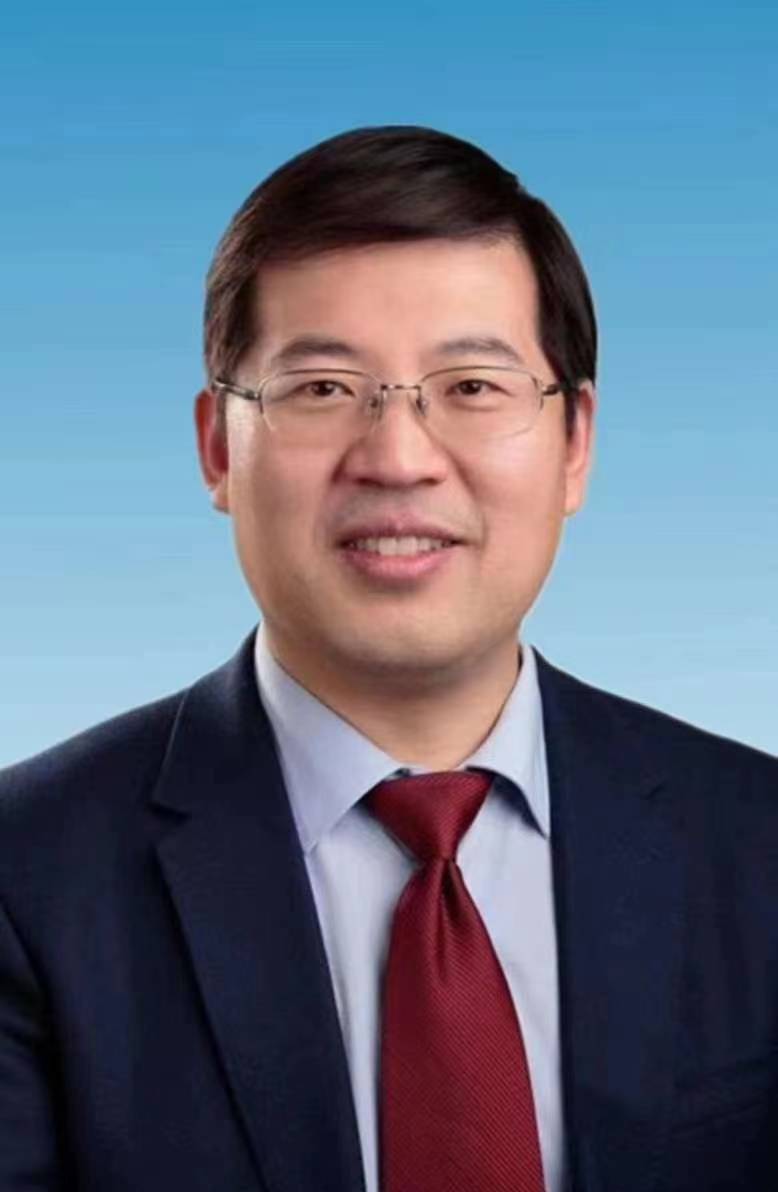}}]{Xinbo Gao}
received the B.Eng., M.Sc. and Ph.D. degrees in electronic engineering, signal and information processing from Xidian University, Xi'an, China, in 1994, 1997, and 1999, respectively. From 1997 to 1998, he was a research fellow at the Department of Computer Science, Shizuoka University, Shizuoka, Japan. From 2000 to 2001, he was a post-doctoral research fellow at the Department of Information Engineering, the Chinese University of Hong Kong, Hong Kong. Since 2001, he has been at the School of Electronic Engineering, Xidian University. He is currently a Cheung Kong Professor of Ministry of Education of P. R. China, a Professor of Pattern Recognition and Intelligent System of Xidian University and a Professor of Computer Science and Technology of Chongqing University of Posts and Telecommunications. His current research interests include Image processing, computer vision, multimedia analysis, machine learning and pattern recognition. He has published six books and around 300 technical articles in refereed journals and proceedings. Prof. Gao is on the Editorial Boards of several journals, including Signal Processing (Elsevier) and Neurocomputing (Elsevier). He served as the General Chair/Co-Chair, Program Committee Chair/Co-Chair, or PC Member for around 30 major international conferences. He is a Fellow of the Institute of Engineering and Technology and a Fellow of the Chinese Institute of Electronics.
\end{IEEEbiography}






\end{document}